\documentclass[manuscript,screen]{acmart}
\AtBeginDocument{%
  }

\setcopyright{acmlicensed}
\copyrightyear{2026}
\acmYear{2026}
\acmDOI{XXXXXXX.XXXXXXX}
\acmISBN{978-1-4503-XXXX-X/2026/03}

\usepackage{acronym}
\usepackage{adjustbox}
\usepackage{tabularx}
\usepackage{booktabs}
\usepackage{verbatim}
\usepackage{array}
\usepackage{textcomp}

\usepackage{hyperref} 
\usepackage{subfigure}
\usepackage{multirow}
\usepackage{subcaption} 
\usepackage{float}
\begin{document}

\newacro{AUC}{Area Under the Curve}
\newacro{BF}{Bloom Filter}
\newacro{CDP}{Central Differential Privacy}
\newacro{DL}{Deep Learning}
\newacro{DML}{Distributed Machine Learning}
\newacro{DNN}{Deep Neural Network}
\newacro{DP}{Differential Privacy}
\newacro{DP-BF}{Differentially Private Bloom Filter}
\newacro{DP-GAN}{Differentially Private Generative Adversarial Network}
\newacro{DP-SGD}{Differentially Private Stochastic Gradient Descent}
\newacro{DTW}{Dynamic Time Warping}
\newacro{ECL}{Electricity Consumption and Occupancy}
\newacro{FGSM}{Fast Gradient Sign Method}
\newacro{FedGAN}{Federated Generative Adversarial Network}
\newacro{FHE}{Fully Homomorphic Encryption}
\newacro{FL}{Federated Learning}
\newacro{FTL}{Federated Transfer Learning}
\newacro{GAN}{Generative Adversarial Network}
\newacro{GC}{Garbled Circuits}
\newacro{GDPR}{General Data Protection Regulation}
\newacro{HE}{Homomorphic Encryption}
\newacro{HIPAA}{Health Insurance Portability and Accountability Act}
\newacro{IIoT}{Industrial IoT}
\newacro{IoT}{Internet of Things}
\newacro{KL}{Kullback-Leibler}
\newacro{LDP}{Local Differential Privacy}
\newacro{LFW}{Labeled Faces in the Wild}
\newacro{LLM}{Large Language Model}
\newacro{LPWAN}{Low-Power Wide-Area Network}
\newacro{MAE}{Mean Absolute Error}
\newacro{MIA}{Membership Inference Attack}
\newacro{ML}{Machine Learning}
\newacro{MMD}{Maximum Mean Discrepancy}
\newacro{non-IID}{non-Independent and Identically Distributed}
\newacro{NREL}{National Renewable Energy Laboratory}
\newacro{PIPL}{Personal Information Protection Law}
\newacro{PPML}{Privacy-Preserving Machine Learning}
\newacro{PPRL}{Privacy-Preserving Record Linkage}
\newacro{PQ}{Post-Quantum}
\newacro{PQ-DP}{Post-Quantum Differential Privacy}
\newacro{QML}{Quantum Machine Learning}
\newacro{RMSE}{Root Mean Squared Error}
\newacro{ROC}{Receiver Operating Characteristic}
\newacro{SGD}{Stochastic Gradient Descent}
\newacro{SL}{Split Learning}
\newacro{SM}{Smart Meter}
\newacro{SMPC}{Secure Multi-Party Computation}
\newacro{TEE}{Trusted Execution Environment}
\newacro{TSTR}{Train on Synthetic, Test on Real}
\newacro{VAE}{Variational Auto-Encoder}

\title{Privacy-Preserving Machine Learning for IoT: A Cross-Paradigm Survey and Future Roadmap}

\author{Zakia Zaman}
\affiliation{%
  \institution{School of CSE, University of NSW}
  \city{Sydney}
  \state{NSW}
  \country{AU}}
\email{z.zaman@unsw.edu.au}

\author{Praveen Gauravaram}
\affiliation{%
  \institution{TCS Research \& Innovation, Tata Consultancy Services (Australia \& New Zealand)}
  \city{Brisbane}
  \state{QLD}
  \country{Australia}
}
\email{p.gauravaram@tcs.com}

\author{Mahbub Hassan}
\affiliation{%
  \institution{School of CSE, University of NSW}
  \city{Sydney}
  \state{NSW}
  \country{AU}}
\email{mahbub.hassan@unsw.edu.au}

\author{Sanjay Jha}
\affiliation{%
  \institution{School of CSE, University of NSW}
  \city{Sydney}
  \state{NSW}
  \country{AU}}
\email{sanjay.jha@unsw.edu.au}

\author{Wen Hu}
\affiliation{%
  \institution{School of CSE, University of NSW}
  \city{Sydney}
  \state{NSW}
  \country{AU}}
\email{wen.hu@unsw.edu.au}

\renewcommand{\shortauthors}{Zaman et al.}

\begin{abstract}
The rapid proliferation of \ac{IoT} has intensified the need for \ac{PPML} mechanisms capable of protecting sensitive data generated by large-scale, heterogeneous, and resource-constrained devices. Unlike centralised learning settings, \ac{IoT} systems are decentralised, bandwidth-limited, latency-sensitive, and vulnerable to privacy risks across sensing, communication, and distributed training pipelines, limiting the effectiveness of traditional protection methods. This survey provides a comprehensive, \ac{IoT}-centric and cross-paradigm review of \ac{PPML}, introducing a structured taxonomy that spans perturbation-based techniques such as \ac{DP}, distributed learning approaches like \ac{FL}, cryptographic methods including \ac{HE} and \ac{SMPC}, and generative models such as \acp{GAN}.
For each paradigm, we analyse privacy guarantees, computational and communication overheads, scalability under device heterogeneity, and robustness against attacks including \ac{MIA}s, model inversion, gradient leakage, and adversarial manipulation. We further examine deployment constraints in wireless and distributed \ac{IoT} environments, emphasising trade-offs among privacy, utility, convergence, communication cost, and energy efficiency, particularly in emerging 5G/6G architectures. Finally, we review evaluation practices, datasets, and frameworks, and outline open challenges, including hybrid designs, energy-aware \ac{FL}, and scalable, trustworthy \ac{PPML} solutions.
\end{abstract}




\maketitle

\section{Introduction}
The Internet of Things (\ac{IoT}) is increasingly embedded in modern life, enabling transformative applications across healthcare, smart homes, industrial automation, smart grids, transportation, and energy management. These interconnected ecosystems rely on continuous sensing, data collection, communication, and analytics, often involving highly sensitive information such as biometric signals, audio streams, geolocation traces, behavioural patterns, and fine-grained energy usage profiles. Protecting the privacy of such data is therefore both a societal necessity and a regulatory imperative, reinforced by legal and policy frameworks such as the \ac{GDPR} \cite{voigt2017eu}, the \ac{HIPAA} \cite{annas2003hipaa}, and China's \ac{PIPL} \cite{calzada2022citizens}.

Privacy preservation in \ac{IoT} environments remains particularly challenging. Unlike traditional centralised computing architectures, \ac{IoT} deployments are inherently decentralised, heterogeneous, and resource-constrained. Devices often operate under strict limits on computation, memory, battery life, and communication bandwidth, while many applications require near-real-time inference and decision-making. These characteristics constrain the direct adoption of conventional privacy-preserving mechanisms and intensify the trade-off between privacy guarantees, model utility, latency, and system efficiency. At the same time, the growing use of edge intelligence and collaborative learning broadens the threat surface, exposing both data and models to attacks such as membership inference, model inversion, gradient leakage, data poisoning, and adversarial perturbation.

Beyond algorithmic considerations, the communication substrate itself strongly influences privacy outcomes in \ac{IoT} systems. Bandwidth-constrained protocols such as \acp{LPWAN}, over-the-air aggregation under noisy or fading wireless channels, and traffic-analysis leakage from encrypted model-update exchanges all require privacy mechanisms to be designed in conjunction with the underlying network layer. This cross-layer dependency fundamentally distinguishes \ac{PPML} in \ac{IoT} from its data-centre counterpart.

\ac{PPML} has emerged as a promising direction for embedding privacy protection directly into the \ac{ML} pipeline. Core \ac{PPML} paradigms include perturbation-based mechanisms such as \ac{DP}, distributed learning approaches such as \ac{FL} \cite{dritsas2025federated}, cryptographic methods including \ac{HE} and \ac{SMPC}, and generative models for privacy-aware data synthesis. While these paradigms have each received substantial attention, their practical adaptation to \ac{IoT} environments introduces additional challenges related to scalability, adversarial robustness, communication overhead, deployment feasibility, and regulatory compliance.

This survey presents a unified and \ac{IoT}-centric analysis of \ac{PPML} methods, bridging fragmented research across perturbation-based approaches, distributed learning, cryptographic computation, and synthetic data generation. We develop a structured taxonomy, map major privacy threats to corresponding defence mechanisms, consolidate evaluation methodologies, and identify open research gaps. In contrast to many recent surveys, this work adopts an explicitly IoT-first and cross-paradigm perspective, with particular emphasis on lightweight, scalable, and practically deployable privacy solutions for adversarially exposed and resource-constrained \ac{IoT} environments.

\begin{table*}[!htbp]
\centering
\caption{List of Acronyms}
\label{tab:acronyms}
\small
\begin{minipage}{0.48\textwidth}
\centering
\begin{tabularx}{\textwidth}{@{} l X @{}}
\toprule
\textbf{Acronym} & \textbf{Expansion} \\
\midrule
AUC    & Area Under the Curve \\
BF     & Bloom Filter \\
CDP    & Central Differential Privacy \\
DL     & Deep Learning \\
DML    & Distributed Machine Learning \\
DNN    & Deep Neural Network \\
DP     & Differential Privacy \\
DP-BF  & Differentially Private Bloom Filter \\
DP-GAN & Differentially Private Generative Adversarial Network \\
DP-SGD & Differentially Private Stochastic Gradient Descent \\
DTW    & Dynamic Time Warping \\
ECL    & Electricity Consumption and Occupancy \\
FGSM   & Fast Gradient Sign Method \\
FedGAN & Federated Generative Adversarial Network \\
FHE    & Fully Homomorphic Encryption \\
FL     & Federated Learning \\
FTL    & Federated Transfer Learning \\
GAN    & Generative Adversarial Network \\
GC     & Garbled Circuits \\
GDPR   & General Data Protection Regulation \\
HE     & Homomorphic Encryption \\
HIPAA  & Health Insurance Portability and Accountability Act \\
IIoT   & Industrial IoT \\
IoT    & Internet of Things \\
\bottomrule
\end{tabularx}
\end{minipage}%
\hfill
\begin{minipage}{0.48\textwidth}
\centering
\begin{tabularx}{\textwidth}{@{} l X @{}}
\toprule
\textbf{Acronym} & \textbf{Expansion} \\
\midrule
KL     & Kullback-Leibler \\
LDP     & Local Differential Privacy \\
LFW     & Labeled Faces in the Wild \\
LLM     & Large Language Model \\
LPWAN   & Low-Power Wide-Area Network \\
MAE     & Mean Absolute Error \\
MIA     & Membership Inference Attack \\
ML      & Machine Learning \\
MMD     & Maximum Mean Discrepancy \\
non-IID & non-Independent and Identically Distributed \\
NREL    & National Renewable Energy Laboratory \\
PIPL    & Personal Information Protection Law \\
PPML    & Privacy-Preserving Machine Learning \\
PPRL    & Privacy-Preserving Record Linkage \\
PQ      & Post-Quantum \\
PQ-DP   & Post-Quantum Differential Privacy \\
QML     & Quantum Machine Learning \\
RMSE    & Root Mean Squared Error \\
ROC     & Receiver Operating Characteristic \\
SGD     & Stochastic Gradient Descent \\
SL      & Split Learning \\
SM      & Smart Meter \\
SMPC    & Secure Multi-Party Computation \\
TEE     & Trusted Execution Environment \\
TSTR    & Train on Synthetic, Test on Real \\
VAE     & Variational Auto-Encoder \\
\bottomrule
\end{tabularx}
\end{minipage}
\end{table*}

\subsection{How This Survey Differs}

Existing surveys on privacy in \ac{IoT} generally fall into three categories: domain-specific reviews centred on sectors such as \ac{IIoT} or smart grids; technique-specific surveys focused on a single \ac{PPML} family such as \ac{FL}, \ac{DP}, or cryptographic computation; and broad security and privacy overviews that treat privacy only at a high level. Representative examples include works on \ac{IIoT}~\cite{jiang2021differential}, smart grids~\cite{iqbal2025privacy}, general \ac{IoT} privacy~\cite{amiri2020survey, al2020survey, wakili2025privacy, aouedi2024survey, abbas2022safety, le2024applications}, \ac{FL}~\cite{raghav2025survey}, and cryptographic approaches~\cite{chen2025privacy}. While these studies provide valuable foundations, they share a common limitation: \ac{PPML} paradigms are treated in isolation, cross-paradigm comparison is limited, and the deployment realities of heterogeneous, bandwidth-constrained, and energy-limited \ac{IoT} systems receive insufficient attention. No existing survey provides a unified analytical framework that evaluates major \ac{PPML} paradigms jointly across privacy guarantees, computational and communication complexity, attack resilience, evaluation practices, datasets, implementation toolchains, and practical deployability.
This survey addresses that gap through an \ac{IoT}-first, cross-paradigm lens, organising the literature into four interrelated \ac{PPML} categories: \textit{(i)}~perturbation-based mechanisms, \textit{(ii)}~distributed learning paradigms, \textit{(iii)}~cryptographic protocols, and \textit{(iv)}~generative privacy models. This taxonomy enables systematic comparison of techniques that are frequently discussed in isolation but increasingly combined in practice. Within it, we emphasise deployability under \ac{IoT} constraints---covering lightweight approaches such as \ac{BF}-based encoding variants and other low-overhead mechanisms---while connecting adversarial threat models to their corresponding defences, consolidating representative datasets and open-source toolchains, and surveying emerging directions including privacy-preserving foundation models, Transformer-based privacy-aware forecasting, hybrid \ac{DP}-cryptographic architectures, and quantum-resilient \ac{PPML}.
Table~\ref{tab:related_surveys} benchmarks representative prior surveys against the present work across scope, contributions, and limitations. Three differentiators stand out: this survey \textit{(i)}~unifies major \ac{PPML} paradigms within a single analytical framework rather than treating any one family in isolation; \textit{(ii)}~grounds every discussion in concrete \ac{IoT} deployment constraints, including communication overhead, energy efficiency, edge execution, and system heterogeneity; and \textit{(iii)}~moves beyond descriptive summarisation to offer a structured threat--defence mapping, a consolidated evaluation landscape, and a forward-looking research roadmap. Accordingly, this work is positioned not as a literature overview, but as a comparative and integrative reference for understanding how diverse \ac{PPML} techniques can be assessed, combined, and adapted for real-world \ac{IoT} ecosystems.

\begin{table*}[htbp!]
\centering
\caption{Summary of representative surveys related to privacy-preserving machine learning in IoT, organised year-wise.}
\label{tab:related_surveys}
\small
\begin{tabular}{|p{3.0cm}|p{3.2cm}| p{4.0cm}| p{4.2cm}|}
\toprule
\textbf{Reference} & \textbf{Focus/Scope} & \textbf{Main Contributions} & \textbf{Limitations / Gaps} \\
\midrule

Amiri-Zarandi \textit{et al.}, 2020 \cite{amiri2020survey} &
\ac{ML}-based privacy in \ac{IoT} (general) &
Early survey linking \ac{ML} techniques with \ac{IoT} privacy; broad overview of solutions. &
Does not systematise formal \ac{PPML} paradigms (\ac{DP}, \ac{FL}, \ac{HE}, synthetic); lacks attack–defence, datasets, and frameworks. \\
\hline

Al-Garadi \textit{et al.}, 2020 \cite{al2020survey} &
\ac{ML}/\ac{DL} for \ac{IoT} security &
Extensive review of \ac{ML}/\ac{DL} in \ac{IoT} security context. &
Security-centric; privacy-preserving aspects only briefly covered. \\

\hline

Jiang \textit{et al.}, 2021 \cite{jiang2021differential} &
Differential Privacy in Industrial IoT (IIoT) &
Detailed review of \ac{DP} techniques for \ac{IIoT}. &
Domain-specific (\ac{IIoT}); excludes broader \ac{IoT} contexts, other paradigms, and datasets. \\

\hline

Abbas \textit{et al.}, 2022 \cite{abbas2022safety} &
Safety, security, privacy in \ac{ML}-based \ac{IoT} &
General overview of safety, security, and privacy concerns in \ac{ML}-enabled \ac{IoT}. &
Not a formal \ac{PPML} survey; lacks taxonomy, evaluation metrics, or datasets/frameworks. \\

\hline

Aouedi \textit{et al.}, 2024 \cite{aouedi2024survey} &
Intelligent \ac{IoT}: applications, security, privacy &
Comprehensive overview of \ac{IoT} applications including security/privacy issues. &
Very broad scope; \ac{PPML} methods treated only at high level. \\

\hline

Le \textit{et al.}, 2024 \cite{le2024applications} &
Distributed \ac{ML} in \ac{IoT} &
Covers architectures and applications of distributed \ac{ML} in \ac{IoT}. &
Distributed \ac{ML} focus; privacy-preserving aspects underexplored. \\
\hline
Wakili \textit{et al.}, 2025 \cite{wakili2025privacy},
Kaur \textit{et al.}, 2025 \cite{Kaur2025IoTPrivacySurvey} &
\ac{IoT} network security and privacy &
Comparative analysis of \ac{IoT} privacy-preserving methods and applications. &
Broad \ac{IoT} orientation; not \ac{PPML}-focused, no taxonomy of learning-based approaches. \\
\hline
Chen \textit{et al.}, 2025 \cite{chen2025privacy} &
Cryptography-based \ac{PPML} &
Comprehensive survey of \ac{HE}, \ac{SMPC}, and related cryptographic \ac{ML} methods. &
Technique-specific; excludes \ac{DP}, \ac{FL}, synthetic data, and \ac{IoT} resource constraints. \\
\hline
Demelius \textit{et al.}, 2025 \cite{demelius2025recent} &
\ac{DP} in centralised deep learning &
Systematic survey of \ac{DP} mechanisms in centralised \ac{DL}; covers \ac{DP-SGD} variants, privacy accounting, and auditing methods. &
Centralised \ac{DL} only; excludes \ac{IoT} deployment constraints, \ac{FL}, cryptographic approaches, and generative models. \\
\hline
Raghav \textit{et al.}, 2025 \cite{raghav2025survey} &
Federated learning for \ac{IoT} &
Targeted survey of privacy-preserving \ac{FL} in \ac{IoT}. &
Narrow (\ac{FL} only); lacks cross-paradigm comparison. \\
\hline
\textbf{This Work (2026)} &
Comprehensive \ac{PPML}--\ac{IoT} Survey &
Unified taxonomy covering \ac{DP}, \ac{FL}, \ac{HE}, \ac{DP-BF}-Encoding, \ac{GAN}s, and \ac{LLM}s; mapping of attacks--defences; systematic dataset and framework review; year-wise trend analysis. &
Bridges prior gaps by integrating all paradigms, including lightweight \ac{DP-BF}-Encoding for \ac{IoT}; positions future directions (quantum-safe \ac{PPML}, 6G \ac{IoT}, energy-aware \ac{PPML}). \\
\hline
\end{tabular}
\vspace{0.2cm}

\footnotesize{\textit{Note:} DP-BF (Differentially Private Bloom Filter) refers to lightweight probabilistic encoding schemes that integrate noise for privacy while retaining utility, making them well-suited for constrained \ac{IoT} environments vulnerable to inference and frequency-based attacks.}
\end{table*}

\subsection{Paper Selection Methodology}
This survey reviews more than 200 publications spanning 2014 to early 2026, including peer-reviewed journal articles, conference papers, arXiv preprints, and selected technical reports. A limited number of foundational works published before 2014 are also included where necessary to establish theoretical context. Literature was collected from major scholarly sources, including the ACM Digital Library, IEEE Xplore, SpringerLink, arXiv, USENIX, leading machine learning venues such as NeurIPS and ICML, and specialised privacy venues such as PETs/PoPETs.

The literature search combined keywords including \textit{``privacy-preserving machine learning''}, \textit{``differential privacy''}, \textit{``federated learning''}, \textit{``homomorphic encryption''}, \textit{``secure multi-party computation''}, \textit{``generative adversarial network''}, \textit{``synthetic data generation''}, \textit{``IoT privacy''}, \textit{``membership inference attack''}, \textit{``gradient leakage''}, and \textit{``Bloom Filter encoding''}, using Boolean combinations where appropriate (for example, \textit{``federated learning AND differential privacy AND IoT''}).

Papers were included if they satisfied one or more of the following criteria:
\begin{enumerate}
    \item they proposed or evaluated \ac{PPML} methods in \ac{IoT} or distributed learning settings;
    \item they analysed privacy--utility trade-offs formally or empirically; or
    \item they addressed system-level constraints such as latency, communication cost, energy efficiency, or regulatory compliance.
\end{enumerate}
Preprints and technical reports were retained where they represented influential or emerging contributions not yet comprehensively covered in archival literature. Papers focused on privacy in non-\ac{ML} contexts or lacking either formal analysis or meaningful empirical evaluation were excluded. The selected literature was then organised thematically across \ac{DP}, \ac{DP-BF}-Encoding, \ac{FL}, cryptographic approaches, synthetic data generation, adversarial threats, evaluation metrics, datasets, and implementation frameworks.

\subsection{Trends in PPML-IoT Research}
As illustrated in Fig.~\ref{fig:ppml_research_trend}, research on \ac{PPML} for \ac{IoT} has grown steadily between 2019 and 2025. The field has evolved from largely conceptual and theoretical studies toward increasingly practical and system-oriented solutions, particularly in areas such as \ac{FL} and \ac{DP}. At the same time, progress in cryptographic computation and synthetic data generation has broadened the design space for privacy-aware \ac{IoT} intelligence.

More recently, privacy-preserving foundation models and Transformer-based architectures have begun to attract attention for distributed inference, secure forecasting, and privacy-aware time-series analysis in \ac{IoT} settings. These developments reflect a broader shift in the literature from isolated privacy mechanisms toward deployable, efficient, and robust \ac{PPML} pipelines that can operate under realistic \ac{IoT} constraints. The observed growth trend also indicates increasing recognition that privacy in \ac{IoT} is not a peripheral concern, but a foundational requirement for trustworthy and scalable intelligent systems.

\subsection{Contributions}
This survey provides a comprehensive and structured investigation of \ac{PPML} techniques tailored to the unique demands of \ac{IoT} systems. The key contributions of this work are as follows:
\begin{itemize}
    \item \textit{Unified cross-paradigm taxonomy:} We introduce an \ac{IoT}-centric taxonomy that organises \ac{PPML} methods across perturbation-based mechanisms, distributed learning paradigms, cryptographic protocols, and generative privacy models. This taxonomy provides a common analytical basis for comparing approaches that are usually studied in isolation.

    \item \textit{Comparative analysis across technical and system dimensions:} We systematically compare \ac{PPML} paradigms in terms of privacy guarantees, attack resilience, utility preservation, computational overhead, communication cost, scalability, and suitability for deployment on heterogeneous and resource-constrained \ac{IoT} devices.

    \item \textit{Threat--defence alignment:} We map major privacy threats, including membership inference, model inversion, gradient leakage, reconstruction, poisoning, and adversarial manipulation, to the defence mechanisms most commonly used to mitigate them in \ac{IoT} settings, thereby clarifying both protection boundaries and residual risks.

    \item \textit{IoT deployment perspective:} We explicitly examine how \ac{PPML} mechanisms interact with practical \ac{IoT} constraints such as limited bandwidth, edge execution, wireless communication characteristics, energy efficiency, and low-latency requirements, which are often underemphasised in general \ac{PPML} surveys.

    \item \textit{Consolidation of evaluation ecosystem:} We synthesise commonly used evaluation dimensions spanning privacy, utility, robustness, and system performance, and complement this with a curated overview of representative datasets, benchmarks, and open-source frameworks relevant to \ac{PPML} in \ac{IoT}.

    \item \textit{Coverage of emerging directions and future roadmap:} In addition to established paradigms, we discuss lightweight and emerging directions such as \ac{DP-BF}-Encoding, privacy-preserving foundation models, hybrid \ac{DP}--cryptographic designs, multimodal privacy-aware learning, energy-aware \ac{PPML}, and quantum-resilient privacy mechanisms, and we identify promising research opportunities for next-generation \ac{IoT} systems.
\end{itemize}
Collectively, these contributions position this survey as a comparative, integrative, and forward-looking reference for researchers and practitioners seeking to design scalable, trustworthy, and regulation-aware \ac{PPML} solutions across diverse \ac{IoT} domains.

\subsection{Organisation of the paper}
The remainder of this paper is structured as follows. Section~\ref{sec:background} provides foundational background on \ac{IoT}–\ac{ML} systems, illustrates privacy exposure risks, and motivates the need for \ac{PPML}. Section~\ref{sec:data_privacy} reviews traditional data privacy models, and Section~\ref{sec:core_ppml} presents the main \ac{PPML} techniques with adaptations for \ac{IoT}. Section~\ref{sec:ppml_attack} explores privacy attack vectors and defences. Section~\ref{sec:evaluation_metric} outlines evaluation metrics for privacy, utility, and robustness, followed by a discussion of benchmark datasets in Section~\ref{sec:datasets_lit}. Section~\ref{sec:ppml_framework} surveys existing implementation frameworks, and Section~\ref{sec:future} identifies research gaps, future directions and lessons learned. Finally, Section~\ref{sec:conclusion} concludes the paper.

\begin{figure}[h]
    \raggedright
    \subfigure[Growth in research publications on \ac{PPML} for \ac{IoT} between 2019 and 2025. The trend highlights steady expansion through 2019–2021, followed by accelerated growth after 2022, driven by \ac{FL}, \ac{DP}, cryptographic enhancements, and more recently \ac{GAN}-based approaches.]{
        \includegraphics[scale=0.25]{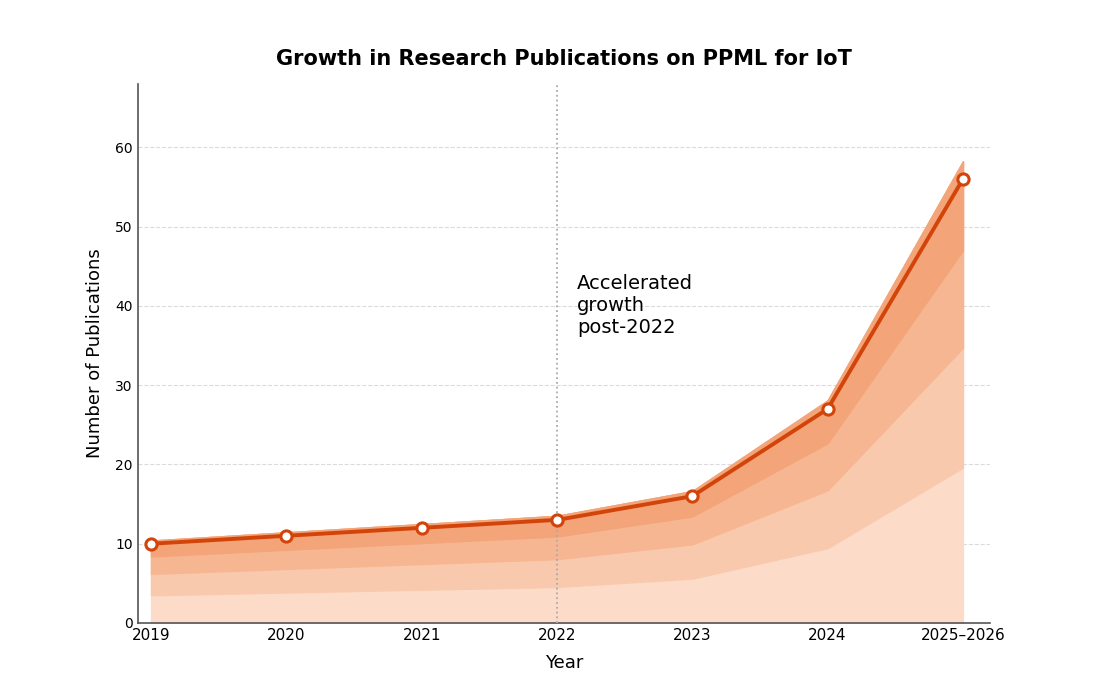}
        \label{fig:ppml_research_trend}
    }
    \hfill
    \subfigure[Representative smart metering system: household sensors stream fine-grained energy consumption data to gateways and cloud services for real-time forecasting, optimisation, and billing. While enabling intelligent energy management, these data pipelines may reveal sensitive behavioural patterns (e.g., occupancy schedules, appliance usage).]{
        \includegraphics[scale=0.25]{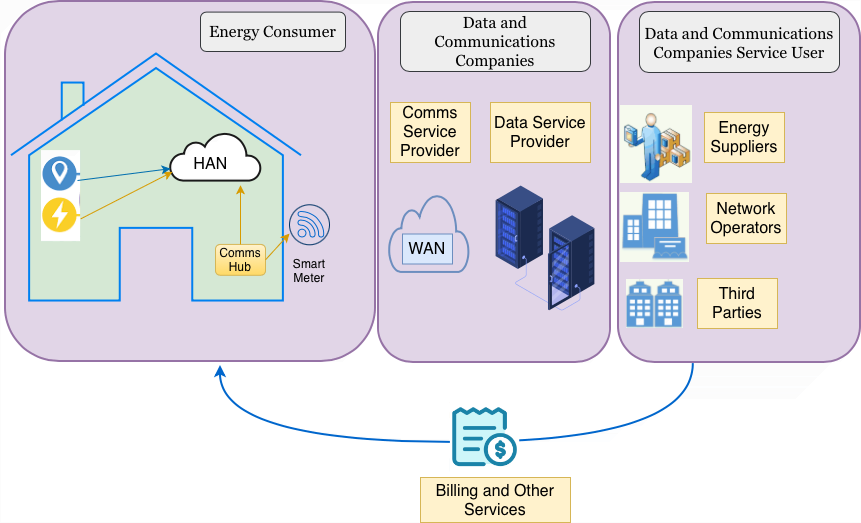}
        \label{fig:iot}
    }
    \caption{(a) Research publication growth on \ac{PPML}. (b) Representative smart meter data flow diagram}
    \label{fig:fig1}
\end{figure}

\begin{figure*}[htbp!]
    \centering
    \includegraphics[scale=.28]{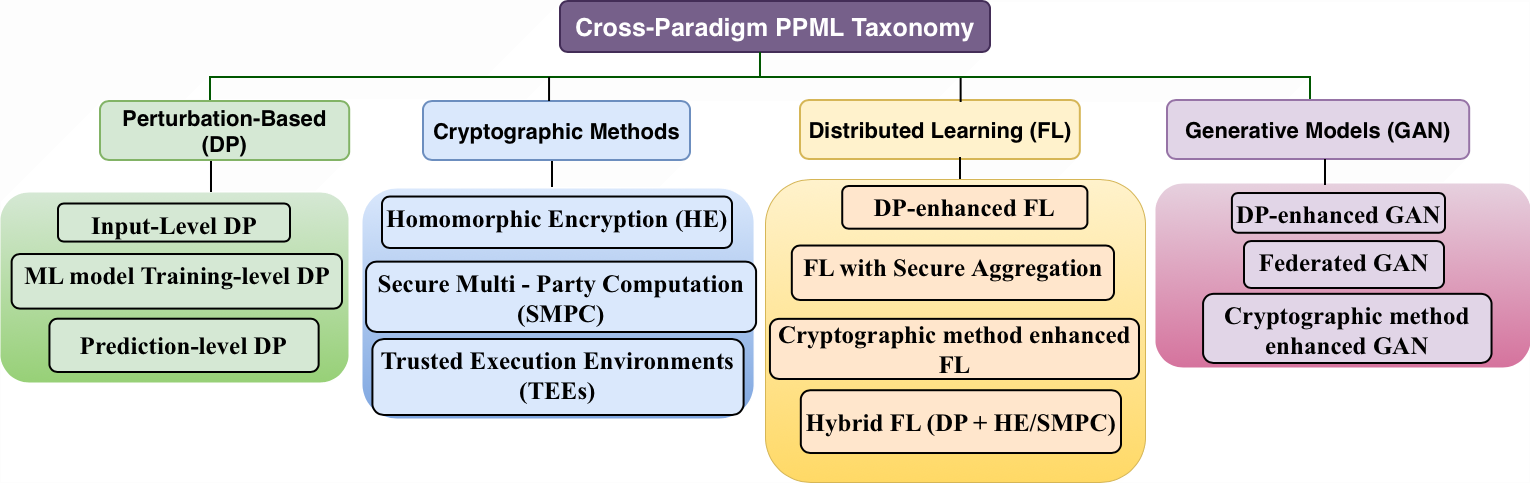}
    \caption{Cross-Paradigm taxonomy of Privacy-Preserving Machine Learning (PPML) techniques for IoT.}  
    \label{fig:taxonomy}
\end{figure*}

\begin{table*}[ht]
\centering
\caption{Characteristics of PPML Paradigm}
\label{tab:ppml_paradigm_comparison}
\renewcommand{\arraystretch}{1.2}
\begin{adjustbox}{max width=\textwidth}
\large
\begin{tabular}{>{\raggedright\arraybackslash}p{2.8cm}|
                >{\raggedright\arraybackslash}p{4.3cm}|
                >{\raggedright\arraybackslash}p{3.2cm}|
                >{\raggedright\arraybackslash}p{3.2cm}|
                >{\raggedright\arraybackslash}p{3.5cm}}
\hline
Paradigm & Core Techniques & Privacy Guarantee & IoT Suitability & Key Threats Addressed \\
\hline

Perturbation-Based (Differential Privacy) &
Laplace and Gaussian mechanisms; Local Differential Privacy (LDP); DP-Bloom Filter encoding; \ac{DP}-\ac{SGD} with gradient clipping and calibrated noise injection. &
Formal $(\varepsilon,\delta)$-differential privacy with quantifiable privacy budget. &
High — lightweight mechanisms suitable for edge devices and resource-constrained IoT nodes. &
Membership inference; gradient leakage; attribute inference; reconstruction attacks. \\
\hline

Distributed Learning (Federated Learning) &
Federated Averaging (FedAvg); DP-FL; secure aggregation; federated transfer learning; SplitFed architectures. &
Statistical protection through decentralized training, optionally strengthened with differential privacy. &
High — raw data remains localized on devices or gateways, reducing centralized exposure. &
Gradient leakage; model inversion; poisoning attacks; malicious client manipulation. \\
\hline

Cryptographic Protocols &
Homomorphic encryption (Paillier, FHE); secure multi-party computation (garbled circuits, secret sharing); trusted execution environments (e.g., Intel SGX). &
Cryptographic or hardware-enforced confidentiality guarantees independent of statistical assumptions. &
Low–Medium — substantial computational and communication overhead in constrained IoT environments. &
Data reconstruction; membership inference; eavesdropping; inference from encrypted computation outputs. \\
\hline

Generative Models (GAN-Based Approaches) &
Conditional GAN (cGAN); FedGAN; DP-GAN; PATE-GAN; VAE-GAN; Transformer-based generative architectures. &
Informal privacy unless explicitly integrated with differential privacy or teacher-student aggregation mechanisms. &
Medium — computationally intensive training; inference comparatively lightweight for deployment. &
Data sharing risks; membership inference via synthetic sample leakage; attribute inference from generated data. \\
\hline

\end{tabular}
\end{adjustbox}
\end{table*}

\subsection{Cross-Paradigm Taxonomy of PPML for IoT}

To provide a coherent analytical foundation, we organise existing \ac{PPML} techniques into four principal families---perturbation-based approaches, distributed learning paradigms, cryptographic methods, and generative model-based techniques---each reflecting a distinct privacy strategy with different deployment trade-offs for resource-constrained \ac{IoT} systems. Table~\ref{tab:ppml_paradigm_comparison} summarises their core techniques, privacy guarantees, \ac{IoT} suitability, and key threats addressed; the taxonomy is further illustrated in Fig.~\ref{fig:taxonomy}.

Perturbation-based approaches inject calibrated statistical noise---via \ac{LDP}, \ac{DP-SGD}, or lightweight encodings such as \ac{DP-BF}---to provide formal $(\varepsilon,\delta)$ guarantees at low computational cost, making them well-suited to edge and embedded devices, though the privacy--utility trade-off requires careful budget calibration. Distributed learning paradigms, exemplified by \ac{FL} variants such as FedAvg, \ac{FTL}, and SplitFed, localise raw data on-device, reducing centralised exposure; however, formal guarantees require complementary mechanisms such as \ac{DP} or secure aggregation, as decentralisation alone remains vulnerable to gradient leakage, model inversion, and poisoning. Cryptographic methods---\ac{HE}, \ac{SMPC}, and \acp{TEE}---offer strong confidentiality independent of statistical assumptions, but their computational and communication overhead constrains real-time feasibility in highly resource-limited deployments. Generative approaches, including \ac{DP-GAN}, PATE-\ac{GAN}, and Transformer-based generative models, replace raw \ac{IoT} data with synthetic representations; while inference is lightweight, training is resource-intensive and privacy guarantees remain informal unless reinforced with \ac{DP} or teacher-student mechanisms.

This taxonomy establishes the comparative basis---across privacy guarantees, threat coverage, and deployment constraints---for the detailed discussion in the sections that follow.

\section{Background}
\label{sec:background}

The rapid convergence of \ac{ML} and the \ac{IoT} has enabled highly personalised, context-aware services across domains such as energy management, healthcare, and industrial automation. However, \ac{IoT} environments are inherently \textit{distributed}, \textit{resource-constrained}, and \textit{heterogeneous}, often relying on untrusted networks and third-party services. This creates a complex privacy landscape in which sensitive data is continuously generated, transmitted, and analysed at scale. This section introduces the privacy challenges in \ac{IoT}-driven \ac{ML} systems, illustrates how breaches can occur, and sets the stage for a systematic review of privacy-preserving solutions. We begin by contextualising \ac{IoT} data pipelines, then highlight the inherent privacy--utility trade-off in \ac{ML}, trace the evolution of privacy techniques, and summarise key paradigms that define the field of \ac{PPML}.

\subsection{IoT System Context and Privacy Exposure}
\label{sec:iot_privacy}

\ac{IoT} systems typically consist of pervasive sensing devices, edge gateways, and cloud-based analytics platforms. Fig.~\ref{fig:iot} illustrates a smart metering workflow, where energy readings are captured at high temporal resolution and transmitted for grid optimisation and dynamic billing. Comparable architectures exist in wearable healthcare, environmental monitoring, and industrial IoT applications.  

While data-driven pipelines in \ac{IoT} ecosystems significantly enhance automation, personalisation, and real-time decision-making across sectors such as healthcare, smart homes, and urban infrastructure, they simultaneously introduce complex privacy risks. These risks arise from continuous data collection, distributed processing, and collaborative learning, each increasing the vulnerability to adversarial exploitation. 
As a result, individuals and organisations face a series of privacy threats, including:

\begin{itemize}
    \item \textit{Linkage Attacks}: Cross-referencing anonymised \ac{IoT} datasets with auxiliary sources (e.g., social media or utility records) can re-identify users \cite{zheng2018data}.
    \item \textit{Inference Attacks}: \ac{ML} models trained on \ac{IoT} data can predict sensitive traits, such as medical conditions or occupancy patterns \cite{chen2020practical}.
    \item \textit{Model Inversion and Gradient Leakage}: In collaborative learning, adversaries can reconstruct raw data or features from model parameters \cite{he2019model} or shared gradients\cite{zhu2019deep}.
    \item \textit{Traffic Analysis}: Even encrypted communication patterns can disclose user activity schedules or device usage trends \cite{msadek2019iot, ding2022adversarial}.
\end{itemize}

These risks demonstrate that conventional anonymisation or encryption alone is insufficient in adversarial and decentralised \ac{IoT} ecosystems.

\subsection{Privacy Attack Taxonomy in Machine Learning}
\label{sec:active_passive}

The growing use of \ac{ML} in \ac{IoT} environments has expanded the attack surface for privacy breaches. Even when raw data is encrypted or anonymised, ML models may leak sensitive information through their outputs, training dynamics, or internal representations. To understand and mitigate these risks, it is essential to classify privacy attacks by their behaviour, timing, and adversarial knowledge \cite{rigaki2023survey}.

\textit{Passive privacy attacks} involve adversaries who observe a model's behaviour without modifying its training or inference process. These attacks are often performed during inference and are commonly referred to as ``honest-but-curious'' threats. By exploiting statistical signals from predictions or confidence scores, attackers can infer training data characteristics or user membership \cite{shokri2017membership, he2019model}. When the adversary additionally has full visibility into model parameters and gradients, a \emph{white-box passive} setting, attacks such as membership inference \cite{shokri2017membership}, model inversion~\cite{he2019model}, and data reconstruction \cite{hitaj2017deep, hilprecht2019monte} become considerably more potent. Effective countermeasures in this setting include \ac{DP} \cite{abadi2016deep}, gradient perturbation \cite{geyer2017differentially}, regularisation to limit overfitting \cite{nasr2018machine}, confidence obfuscation, and restricting the retention of intermediate feature vectors \cite{hilprecht2019monte}.

\textit{Active privacy attacks}, by contrast, involve adversaries who interfere with the ML pipeline. This includes injecting adversarial inputs, modifying queries, or manipulating training data to elicit specific behaviours or disclosures. Active attacks may occur during both training and inference, and are particularly damaging in collaborative or distributed settings \cite{zhu2019deep, tramer2016stealing}. In the \emph{white-box active} case, where the adversary has full access to model internals and can inject updates, threats such as gradient leakage~\cite{zhu2019deep} and data 
poisoning \cite{biggio2012poisoning} are most severe. Defences in this regime include gradient clipping~\cite{sun2019can}, secure aggregation~\cite{bonawitz2017practical}, homomorphic encryption~\cite{gentry2009fully}, and the application of \ac{DP} during 
local training~\cite{abadi2016deep}.

These attacks also vary by the level of knowledge the attacker possesses. \emph{Black-box} adversaries have access only to model outputs, without visibility into internal parameters or gradients. In the \emph{black-box passive} setting, attackers exploit output probabilities or confidence scores to mount output-based membership inference attacks~\cite{shokri2017membership} or property inference attacks~\cite{ganju2018property}; corresponding defences include output perturbation via \ac{DP}~\cite{dwork2014algorithmic}, model distillation~\cite{tramer2016stealing}, and confidence clipping~\cite{salem2018ml}. In the \emph{black-box active} setting, adversaries combine query access with the ability to craft and submit adversarial inputs, enabling model extraction~\cite{tramer2016stealing, jiang2023comprehensive} and adversarial example generation~\cite{szegedy2013intriguing}. Countermeasures in this case include query rate limiting~\cite{juuti2019prada}, adversarial 
training~\cite{madry2017towards}, and input preprocessing~\cite{rahman2018membership}.

\subsection{Navigating the Privacy/Utility Trade-Off in Machine Learning}
\label{sec:privacy_utility}
As \ac{IoT} systems increasingly rely on \ac{ML} for prediction, classification, and control, they depend on high-resolution, real-time data. However, privacy-preserving constraints often necessitate transformations that degrade data quality, directly affecting model utility. 

This tension is not unique to energy systems; in healthcare, wearable device data must balance privacy with diagnostic precision, while in industrial IoT, production optimisation competes with confidentiality requirements. Consequently, designing \ac{PPML} systems becomes a multi-objective optimisation challenge, balancing the competing demands of privacy protection, computational efficiency, and model performance across diverse IoT domains.

\subsection{Emergence of Privacy-Preserving Machine Learning (PPML) Paradigms}
To address the limitations of heuristic privacy techniques and mitigate the trade-offs outlined above, \ac{PPML} has emerged as a promising paradigm that embeds privacy directly into learning and inference processes. Core paradigms include:
\begin{itemize}
    \item \ac{DP}: A formal privacy framework that limits the influence of individual data points by injecting calibrated noise~\cite{dwork}.
    \item \ac{FL}: A distributed paradigm that keeps data local while collaboratively training global models~\cite{mcmahan2017communication}, though still vulnerable to gradient leakage and poisoning.
    \item Cryptographic Techniques: Schemes such as \ac{HE}, \ac{SMPC}, and \ac{TEE} offer strong guarantees but face scalability and performance challenges~\cite{ohrimenko2016oblivious}.
    \item Synthetic Data Generation: Generative models, including \ac{GAN} and its \ac{DP}-enhanced variants~\cite{xie2018differentially, jordon2018pate}, produce realistic data substitutes, enabling analytics and sharing while reducing disclosure risk.
\end{itemize}

Collectively, these paradigms form the foundation of a new generation of privacy-preserving solutions, specifically adapted to the constraints and demands of \ac{IoT} ecosystems. However, each approach presents inherent trade-offs, ranging from limited scalability and computational inefficiency to reduced robustness in adversarial settings. In response, emerging research increasingly advocates for hybrid frameworks that integrate lightweight data encoding, formal privacy guarantees, and secure computation to strike a more practical balance.
Building on this foundation, the following sections provide a structured analysis of traditional data privacy mechanisms (Section~\ref{sec:data_privacy}), core \ac{PPML} techniques (Section~\ref{sec:core_ppml}), adversarial threat models and corresponding defences (Section~\ref{sec:ppml_attack}), and the evaluation metrics used to assess their effectiveness (Section~\ref{sec:evaluation_metric}).

\section{Traditional Data Privacy Paradigms}
\label{sec:data_privacy}
Traditional data privacy mechanisms emerged in response to re-identification risks in shared datasets. Domains such as healthcare, finance, and smart environments relied on anonymisation~\cite{efthymiou2010smart} or probabilistic obfuscation~\cite{dinusha, smith2017secure} to balance utility with confidentiality. However, most techniques were designed for centralised, structured datasets and struggle to meet the formal guarantees and operational constraints of modern \ac{ML} and \ac{IoT} systems. This section reviews two foundational approaches---anonymisation and probabilistic data encoding---highlighting their principles, applications, and limitations in decentralised, high-frequency data environments.

\subsection{Anonymisation}
\label{sec:anonymisation}

Anonymisation reduces identifiability by removing explicit identifiers and perturbing quasi-identifiers. Classical approaches such as $k$-anonymity~\cite{sweeney2002k}, $l$-diversity~\cite{machanavajjhala2007diversity}, and $t$-closeness~\cite{li2006t} define group-based indistinguishability criteria, while imaging methods like $k$-same~\cite{gross2006model, newton2005preserving} average facial features across $k$ individuals to produce non-identifiable synthetic representations.

Despite being computationally lightweight, these techniques exhibit well-documented vulnerabilities: auxiliary information and quasi-identifier correlation can defeat the protection~\cite{narayanan2006break, machanavajjhala2007diversity, lermen2026deanonymization}, generalisation degrades utility in spatial and temporal \ac{IoT} data~\cite{zang2011anonymization, saraswathi2025Location}, and reliance on a trusted central curator introduces a single point of failure~\cite{vergara2016privacy, minhong2025Trajectory}.

\subsection{Probabilistic Data Encoding: Bloom Filters}
\label{sec:probabilistic_bf}

\ac{BF} encode categorical or string-based values into fixed-length bit vectors via multiple hash functions, enabling approximate matching without disclosing raw data~\cite{bloom, rainer, dinusha1}. Their computational efficiency and data minimisation properties make them attractive for privacy-preserving record linkage in distributed environments.

However, \ac{BF} lack formal privacy guarantees and are susceptible to inference attacks. Sensitive attributes such as surnames can be recovered from encodings~\cite{niedermeyer2014cryptanalysis}, and adversarial methods exploiting graph-based co-occurrence structures or bit-pattern frequency distributions can reverse-engineer encoded content~\cite{anushka, schnell2016}. These vulnerabilities are most acute in low-entropy domains where attackers hold background knowledge. Enhancements such as RAPPOR~\cite{ulfer2014} add randomised response to counter frequency-based inference but remain largely unvalidated in practical deployments.

\subsection*{Summary}

Anonymisation and \ac{BF}-based encoding are foundational privacy techniques, yet their heuristic nature and fragility under adversarial, high-dimensional, and decentralised \ac{IoT} conditions limit their adequacy. These shortcomings motivate embedding privacy more formally within the learning and inference process---the focus of the following section on \ac{PPML}.

\section{Core Privacy-Preserving Machine Learning (PPML) Techniques}

\label{sec:core_ppml}
\ac{PPML} has gained considerable momentum in recent years, driven by increasing concerns over the exposure of sensitive data during model training and inference. A range of techniques has been developed to enable collaborative model training across multiple parties while preserving data confidentiality. These approaches primarily draw on cryptographic methods and perturbation-based mechanisms, such as \ac{DP} \cite{rubaie}. The following subsections explore these core techniques, with particular emphasis on their relevance and applicability to \ac{IoT} environments. Special attention is given to \ac{DP}, widely regarded as the gold standard in data privacy, due to its formal privacy guarantees and practical adoption. Each paradigm discussed below corresponds to a primary branch in the cross-paradigm taxonomy (Fig.~\ref{fig:taxonomy}), with hybrid variants mapping to the interconnected sub-nodes.

\subsection{Differential Privacy}
\label{sec:dp_lit}

\ac{DP} offers a mathematically rigorous framework for safeguarding individual privacy by injecting calibrated noise into computations~\cite{demelius2025recent}. Unlike cryptographic methods, \ac{DP} is computationally efficient, making it appealing for large-scale \ac{ML} and resource-constrained \ac{IoT} environments. At its core, \ac{DP} ensures that computation outputs do not significantly differ whether or not any single data point is included, providing strong resistance against inference attacks.

\begin{definition}[$\epsilon$-\ac{DP} \cite{dwork}]
A randomised mechanism $M$ satisfies $\epsilon$-\ac{DP} if for all datasets $D_1$, $D_2$ differing in at most one element and all subsets $S \subseteq \text{Range}(M)$,
\begin{equation}
    \Pr[M(D_1) \in S] \leq \exp(\epsilon) \cdot \Pr[M(D_2) \in S]
\end{equation}
where $\epsilon$ is the privacy budget: smaller values indicate stronger privacy.
\end{definition}

\begin{definition}[$(\epsilon, \delta)$-\ac{DP} \cite{dwork2006our}]
A mechanism $M$ satisfies $(\epsilon, \delta)$-\ac{DP} if for all neighbouring datasets $D_1$, $D_2$ and all $S \subseteq \text{Range}(M)$,
\begin{equation}
    \Pr[M(D_1) \in S] \leq \exp(\epsilon) \cdot \Pr[M(D_2) \in S] + \delta
\end{equation}
where $\delta$ permits a bounded probability of excess leakage; in practice $\delta \ll \frac{1}{n}$ is recommended~\cite{ponomareva2023dp}.
\end{definition}

\subsubsection*{Variants and Mechanisms}

\ac{DP} is deployed in two principal configurations. \textit{\ac{CDP}} relies on a trusted aggregator that collects raw data and adds noise to the output, offering high utility at the cost of centralised trust~\cite{naseri2020local}. \textit{\ac{LDP}} moves noise addition directly onto the user's device before any transmission, eliminating the need for a trusted curator and making it well-suited to \ac{IoT} scenarios with untrusted cloud providers~\cite{mahawaga2022local, zhao2024ScenarioDP}.

Four canonical mechanisms implement these configurations. The \textit{Laplace mechanism} adds noise calibrated to $\ell_1$-sensitivity~\cite{dwork2006calibrating}. The \textit{Gaussian mechanism} adds Gaussian noise to achieve $(\epsilon,\delta)$-\ac{DP}, suited to high-dimensional settings~\cite{abadi2016deep}. The \textit{exponential mechanism} handles non-numeric queries by selecting outputs probabilistically according to a utility function~\cite{mcsherry2007mechanism}. \textit{Randomised response} is a simple survey-derived mechanism still widely used in distributed and \ac{LDP} systems~\cite{ulfer2014, qing2025LDP}.

\subsubsection*{DP in the ML Pipeline}

\ac{DP} can be applied at multiple stages of the \ac{ML} pipeline~\cite{ponomareva2023dp, zheng2025RelaxedDP}. At the \textit{input level}, noise is added to raw data before training, enforcing privacy before data leaves the device---a key requirement in decentralised \ac{IoT} deployments~\cite{feyisetan2020privacy}. At the \textit{training level}, protection can target labels alone~\cite{wu2022does} or the full dataset via mechanisms such as \ac{DP-SGD}~\cite{abadi2016deep}. At the \textit{prediction level}, noise is applied to model outputs such as probability vectors, suitable when a centralised model serves remote clients~\cite{dwork2018privacy}. Despite its strong theoretical appeal, practical deployment of \ac{DP} presents challenges discussed in Section~\ref{sec:future}.


\subsection{Cryptographic Approaches}
\label{sec:cryptographic}

Cryptographic methods enable computation on distributed or encrypted data without exposing sensitive inputs, making them valuable in \ac{IoT} environments where raw data cannot be centralised due to privacy or regulatory constraints. The three dominant paradigms---\ac{HE}, \ac{SMPC}, and \ac{TEE}s---each provide strong formal guarantees but differ in scalability, latency, and trust assumptions.

\textit{\ac{HE}} permits arithmetic operations directly on ciphertexts, keeping intermediate computations private. \ac{FHE} supports arbitrary operations~\cite{gentry2009fully} but at high computational cost; lighter additive schemes such as Paillier are preferred in practice. Applications include privacy-preserving recommender systems~\cite{erkin2012generating} and secure gradient aggregation in \ac{FL}~\cite{yang2024privacy}. However, high overhead and slow encryption cycles make \ac{HE} unsuitable for latency- and energy-sensitive \ac{IoT} deployments.

\textit{\ac{SMPC}} enables multiple parties to jointly evaluate a function without revealing private inputs~\cite{yan2025comet}, using primitives such as \ac{GC}~\cite{yao1986generate} and secret sharing~\cite{shamir1979share}. It has been applied to collaborative \ac{IoT} data sharing, secure distributed regression~\cite{nikolaenko2013privacy}, and \ac{FL} aggregation~\cite{bonawitz2017practical, ma2024trusted}. Its high communication overhead and poor efficiency for non-linear operations, however, limit scalability in bandwidth- and energy-constrained networks.

\textit{\ac{TEE}s} (e.g., Intel SGX, AMD SEV) use hardware-enforced enclaves to isolate sensitive computations from the host OS, supporting confidential \ac{ML} training and inference on untrusted servers~\cite{ohrimenko2016oblivious}. Edge \ac{IoT} frameworks have leveraged \ac{TEE}s for sensor-level protection and enclave-based model execution~\cite{valadares2021trusted}. Nevertheless, limited secure memory, hardware-vendor trust dependencies, and vulnerability to side-channel and speculative execution attacks~\cite{costan2016intel, schneider2022sok} are pronounced concerns in heterogeneous, long-lived \ac{IoT} deployments.

Despite their strong guarantees, all three paradigms face significant challenges in computational efficiency, scalability, and trust assumptions for practical \ac{IoT} use---further discussed in Section~\ref{sec:future}.

\subsection{Federated Learning}
\label{sec:fl}

\ac{FL}~\cite{mcmahan2017communication, shahin2025two, rahmati2025federated, ye2025fedrw} is a \ac{DML} paradigm in which clients collaboratively train a shared global model without exposing raw data. In each round, selected participants train locally and transmit parameter updates to a central server, which aggregates them---typically via weighted averaging (FedAvg)---to update the global model~\cite{Myakala2025PrivacyPre}. While \ac{FL} preserves data locality and reduces centralised storage risks, gradients and model updates can still leak sensitive information~\cite{zhu2019deep}, and local training imposes non-trivial computational and energy overhead on resource-constrained \ac{IoT} nodes.

Several techniques mitigate these risks. \textit{\ac{DP}-enhanced \ac{FL}} injects calibrated noise into local updates to prevent membership or attribute inference~\cite{dwork2018privacy, ranaweera2025federated}. Tang \textit{et al.}~\cite{tang2025FLDP} propose a reputation-based aggregation framework that detects rogue clients injecting excessive noise, preserving model accuracy without sacrificing formal \ac{DP} guarantees. Zhang \textit{et al.}~\cite{zhang2025FLDP} extend this with verifiable \ac{DP} and Byzantine resilience~\cite{xia2025byzantine}, preventing adversarial manipulation of noise injection. \textit{Secure aggregation} protocols~\cite{bonawitz2017practical, wang2025RASA} ensure the server observes only aggregated updates, mitigating targeted inference. \textit{Cryptographic enhancements}---including \ac{HE}~\cite{aono2017privacy} and functional encryption~\cite{boan2025FLfunctionalEncryption}---allow encrypted updates to be aggregated without decryption, extending applicability to resource-constrained devices. \textit{Hybrid approaches} combine \ac{DP}, cryptography, and lightweight compression to jointly reduce privacy risk and communication overhead in \ac{IoT}-\ac{FL} settings~\cite{ma2024trusted}.

Despite its promise, \ac{FL} presents several limitations in \ac{IoT} deployments, discussed in Section~\ref{sec:future}.

\subsection{Synthetic Data by Generative Adversarial Networks (GANs) as a Privacy-Preserving Technique}
\label{sec:synthetic_data}

\ac{GAN} was first introduced by Goodfellow \textit{et al.}~\cite{goodfellow2014generative} as a framework for generating high-quality synthetic data by combining \ac{DNN}s with a game-theoretic minimax optimisation process. The architecture consists of two competing networks: the \textit{Generator}, which maps a random noise distribution into the data space to produce synthetic samples, and the \textit{Discriminator}, which seeks to distinguish real data from synthetic data. Over iterative training, the generator improves to fool the discriminator, thereby approximating the underlying data distribution. Formally, the generator--discriminator game is represented as:
\begin{equation}
    \min _G \max _D \underset{\boldsymbol{x} \sim \mathbb{P}_r}{\mathbb{E}}[\log (D(\boldsymbol{x}))]+\underset{\tilde{\boldsymbol{x}} \sim \mathbb{P}_g}{\mathbb{E}}[\log (1-D(\tilde{\boldsymbol{x}}))]
\end{equation}
where $\mathbb{P}_r$ denotes the real data distribution and $\mathbb{P}_g$ represents the distribution induced by the generator $G$, with $\tilde{\boldsymbol{x}} = G(\boldsymbol{z})$ and $\boldsymbol{z}$ sampled from a noise distribution $p(\boldsymbol{z})$~\cite{gulrajani2017improved}.

Synthetic data generated by \ac{GAN}s has become an increasingly popular privacy-preserving substitute for sensitive datasets, particularly in domains where ethical, legal, or regulatory frameworks (e.g., \ac{GDPR}, \ac{HIPAA}) restrict direct data sharing. By capturing the statistical characteristics of sensitive datasets without reproducing exact individual records, \ac{GAN}s enable safe data release, \ac{PPML}, and algorithm benchmarking. Applications span multiple domains: in healthcare, \ac{GAN}s have generated synthetic patient records, time-series physiological data, and medical images, supporting clinical research without violating patient confidentiality~\cite{yoon2020anonymization, goncalves2020generation, ashrafi2023protect}; in the \ac{IoT} domain, they simulate behavioural patterns, mobility traces, and sensor data, enabling privacy-aware analytics and stress-testing of smart environments.

Newer advances extend vanilla \ac{GAN}s to specialised variants that address privacy--utility trade-offs in distributed and resource-constrained environments. \textit{Conditional GANs (cGANs)} generate synthetic samples conditioned on auxiliary information such as labels or attributes, enabling fine-grained control over data synthesis for categorical or time-series \ac{IoT} tasks---including medical time-series generation and behavioural pattern synthesis in smart homes~\cite{esteban2017real}. Their primary limitations are training instability and limited scalability for high-dimensional \ac{IoT} data. \textit{\ac{FedGAN}s} extend the \ac{GAN} framework to federated settings where generators and discriminators are distributed across devices without centralising raw data, enabling collaborative synthetic \ac{IoT} data generation across sensors or edge nodes~\cite{rasouli2020fedgan, han2025FedGan}. However, this design incurs communication overhead and introduces gradient leakage risks during distributed training.

To provide formal privacy guarantees, \textit{\ac{DP-GAN}s} introduce calibrated noise during training via gradient perturbation~\cite{xie2018differentially}, supporting privacy-aware synthetic \ac{IoT} datasets for anomaly detection and predictive maintenance. While \ac{DP-GAN} provides rigorous $(\epsilon, \delta)$ guarantees, high privacy budgets lead to utility loss and vulnerability to mode collapse. \textit{PATE-GAN}~\cite{jordon2018pate} takes a complementary approach, leveraging teacher-student training under the Private Aggregation of Teacher Ensembles (PATE) framework for safe sample generation in healthcare \ac{IoT} settings where privacy-sensitive attributes must be protected; its main constraint is the requirement for large, disjoint teacher datasets, which limits scalability in \ac{IoT} environments. \textit{VAE-GANs}~\cite{razghandi2022variational} combine Variational Autoencoders with \ac{GAN}s to enhance sample diversity and preserve latent representations, producing synthetic \ac{IoT} sensory data with improved distribution fidelity, though at higher computational cost and with persistent training instability. Finally, \textit{Transformer-based GANs}~\cite{li2022tts, mamat2025emerging} integrate attention mechanisms for stable training and sequence modelling, making them well-suited for smart grid demand forecasting and \ac{IoT} anomaly detection with sequential synthetic data; this direction remains emerging, requiring large-scale data and compute, with robustness not yet well studied.

Across all variants, \ac{GAN}s share common limitations: mode collapse, training instability, and susceptibility to membership inference attacks~\cite{xu2019modeling, zhang2023generative}. Hybrid models such as \ac{DP-GAN} and PATE-GAN mitigate but do not fully eliminate these risks, and privacy--utility trade-offs remain an open research challenge, discussed further in Section~\ref{sec:future}.

\subsection{Overall Comparative Summary}
The Table~\ref{tab:ppml_comparison} presents a consolidated comparison of privacy-preserving approaches, summarising their respective advantages, limitations, computational and communication costs, privacy guarantees, IoT suitability, and representative use-cases, as reported in existing research.

\begin{table*}[htbp!]
\centering
\caption{Privacy-Preserving Techniques in IoT: Comparison of Characteristics, Guarantees, Deployment Trade-offs, and Contrasts with Traditional Methods}
\label{tab:ppml_comparison}
\renewcommand{\arraystretch}{1.2}
\begin{adjustbox}{max width=\textwidth}
\Large
\begin{tabular}{>{\raggedright\arraybackslash}p{2.9cm}|
                >{\raggedright\arraybackslash}p{3.5cm}|
                >{\raggedright\arraybackslash}p{3.5cm}|
                >{\raggedright\arraybackslash}p{2.5cm}|
                >{\raggedright\arraybackslash}p{2.5cm}|
                >{\raggedright\arraybackslash}p{2.0cm}|
                >{\raggedright\arraybackslash}p{2.3cm}|
                >{\raggedright\arraybackslash}p{2.3cm}|
                >{\raggedright\arraybackslash}p{2.9cm}}
\hline
\textbf{Category} &
\textbf{Advantages} &
\textbf{Limitations / Vulnerabilities} &
\textbf{Comp./Comm. Cost} &
\textbf{Privacy Guarantee} &
\textbf{IoT Suitability} &
\textbf{Scalability} &
\textbf{Deployment Status} &
\textbf{Representative Use-Cases} \\
\hline

Anonymisation
\cite{sweeney2002k, machanavajjhala2007diversity, fung2010privacy} &
Simple to implement; well-established in regulations; no special hardware needed; effective for structured, low-dimensional data \cite{fung2010privacy, alwarafy2020survey}. &
Vulnerable to linkage and background-knowledge attacks \cite{narayanan2008robust}; ineffective for high-dimensional/temporal \ac{IoT} data \cite{alwarafy2020survey}; no formal bounds \cite{samarati2002protecting}. &
Low \cite{fung2010privacy} &
Heuristic; no formal guarantees \cite{sweeney2002k} &
Low--Medium \cite{alwarafy2020survey} &
Limited for streaming or high-dimensional data &
Widely deployed but increasingly deprecated &
Census publishing \cite{sweeney2002k}, healthcare records \cite{machanavajjhala2007diversity}, basic \ac{IoT} logs \cite{alwarafy2020survey} \\
\hline

BF Encoding \cite{xue, zaman2024privacy} &
Lightweight, scalable approximate matching; no direct disclosure of raw values; compatible with \ac{PPRL} workflows \cite{rainer, dinusha1}. &
No formal guarantees; susceptible to bit-pattern mining and graph-based inference \cite{anushka}; re-identification risk on low-entropy inputs \cite{niedermeyer2014cryptanalysis}. &
Low &
None; reliant on entropy and attacker ignorance &
Low--Medium &
Efficient but limited to matching and \ac{PPRL} tasks &
Experimental; mostly academic &
Privacy-preserving record linkage (\ac{PPRL}), data matching \cite{dinusha1} \\
\hline

Perturbation-based (DP)
\cite{dwork2014algorithmic, wang2021local, kasiviswanathan2011can} &
Strong formal $(\epsilon,\delta)$ guarantees \cite{dwork2014algorithmic}; lightweight for edge devices \cite{liu2024privacy}; scalable; \ac{LDP} avoids raw data transfer \cite{wang2021local, Qian2025AppliedIntelligenceLDP}. &
Accuracy loss for complex/temporal data \cite{truong2021privacy}; $(\epsilon,\delta)$ tuning trade-offs \cite{abadi2016deep}; membership inference if budget exhausted. &
Low--Moderate \cite{liu2024privacy, truong2021privacy} &
Formal, tunable via $(\epsilon,\delta)$ \cite{dwork2014algorithmic} &
High for low-power \ac{IoT} \cite{truong2021privacy, liu2024privacy} &
Scales to large-scale FL and streaming data \cite{zhao2024privacy} &
Emerging production (e.g., DP in iOS, Chrome) &
Smart meter analytics \cite{hassan2019differential}, healthcare \ac{IoT} \cite{liu2024privacy}, location services \cite{zhao2024privacy, ouatu2025LLM} \\
\hline

Cryptographic-based
\cite{goldreich1998secure, gentry2009fully, costan2016intel} &
Strong theoretical guarantees \cite{goldreich1998secure}; computation on encrypted/partitioned data \cite{acar2018survey}; hardware-assisted security via \ac{TEE} \cite{costan2016intel}. &
High computational/memory cost \cite{gentry2009fully}; real-time \ac{IoT} latency challenges \cite{khan2020lightweight}; \ac{TEE} side-channel risks; vendor trust dependency \cite{costan2016intel}. &
Moderate--Very High \cite{acar2018survey, khan2020lightweight} &
Cryptographic or hardware-enforced \cite{goldreich1998secure, costan2016intel} &
Low--Medium depending on device capability \cite{khan2020lightweight} &
Poor for DL workloads; moderate for lightweight \ac{HE} &
Limited deployment; active research &
Secure smart grid aggregation \cite{acar2018survey}, industrial \ac{IoT} gateways \cite{alwarafy2020survey} \\
\hline

Federated Learning (FL)
\cite{mcmahan2017communication, kairouz2021advances, konevcny2016federated} &
Retains data locality \cite{mcmahan2017communication}; supports heterogeneous devices \cite{kairouz2021advances}; adaptable with compression \cite{shah2021model}; formal privacy with \ac{DP} \cite{geyer2017differentially}. &
Gradient leakage \cite{zhu2019deep}; uneven participation \cite{kairouz2021advances}; \ac{non-IID} convergence degraded under \ac{DP}; poisoning attacks on aggregation. &
Low--Moderate with compression \cite{shah2021model} &
Statistical; strengthened with \ac{DP} \cite{geyer2017differentially} &
High for distributed \ac{IoT} \cite{truong2021privacy, nguyen2021federated} &
Scales with engineering; compression and pruning help &
Emerging production (e.g., FL in Android, Gboard) &
Predictive maintenance \cite{purkayastha2024federated}, mobile sensing \cite{nguyen2021federated}, sensor networks \cite{mcmahan2017communication} \\
\hline

GAN-based
\cite{goodfellow2014generative, yoon2019time, esteban2017real} &
Generates realistic synthetic data \cite{goodfellow2014generative}; preserves temporal patterns \cite{yoon2019time}; avoids raw data sharing \cite{li2022tts}; supports data augmentation. &
Memorisation risk \cite{mukherjee2021privgan}; mode collapse suppresses rare events; informal privacy unless \ac{DP} integrated \cite{torkzadehmahani2019dp}; training instability on edge devices. &
Moderate--High (training); Low (inference) \cite{wen2024generative} &
Informal unless \ac{DP} integrated \cite{torkzadehmahani2019dp} &
Medium--High with compression \cite{wen2024generative} &
Computationally intensive; inference lightweight after training &
Active research; limited production use &
Anomaly detection \cite{lim2024future}, private time-series \cite{yoon2019time}, multimodal \ac{IoT} synthesis \cite{esteban2017real} \\
\hline
\end{tabular}
\end{adjustbox}
\end{table*}


\section{Attacks on Privacy-Preserving Machine Learning (PPML) Systems}
\label{sec:ppml_attack}

Despite their objective of protecting sensitive information, \ac{PPML} systems remain susceptible to a variety of privacy threats that exploit model internals, observable behaviour, and the communication substrate. To systematically examine these threats, we adopt a taxonomy (Sec.~\ref{sec:active_passive}) based on the adversary's capabilities (\textit{white-box} vs.~\textit{black-box}) and intent (\textit{passive} 
vs.~\textit{active})~\cite{hu2022membership, vassilev2024adversarial, niu2024survey}. Table~\ref{tab:ppml_paradigm_comparison} summarises the key threats addressed by each \ac{PPML} paradigm and their primary defences; the following subsections provide a focused analysis of each attack class and its countermeasures.

\subsection{Membership Inference Attack}
\label{sec:mia_lit}

\ac{MIA}s aim to determine whether a given data sample was part of a model's training dataset~\cite{shokri2017membership, hayes2017logan, carlini2021extracting, hu2022membership, wang2025rigging}. In sensitive \ac{IoT} domains such as healthcare and smart metering, even confirming a user's participation in a dataset constitutes a privacy breach. \ac{MIA}s exploit the tendency of \ac{ML} models to behave differently 
on training data (members) versus unseen data (non-members), typically through overfitting. Adversaries leverage these behavioural differences---confidence scores, loss values, and prediction entropy---to infer membership status. \ac{MIA}s are broadly classified into two categories:

\begin{enumerate}
    \item \textit{Black-box attacks}: the adversary has query access only and uses output probabilities or confidence vectors to distinguish members from non-members~\cite{shokri2017membership, yeom2018privacy}.
    \item \textit{White-box attacks}: the attacker has access to model parameters, gradients, or loss functions, enabling more precise 
    inferences~\cite{nasr2019comprehensive}.
\end{enumerate}

To improve attack effectiveness, adversaries employ shadow models trained on surrogate data generated via model inversion ~\cite{he2019model}, statistics-based synthesis ~\cite{carlini2022membership}, or noise-augmented real data ~\cite{truex2019demystifying}.

\noindent\textbf{Defences.} \ac{DP} training injects calibrated noise to reduce membership distinguishability~\cite{abadi2016deep, rahman2018membership}. Regularisation techniques such as dropout and weight decay limit overfitting and associated leakage~\cite{li2020membership}. Prediction sanitisation---returning only class labels or clipping confidence scores---further narrows the adversary's signal~\cite{salem2018ml}.

\subsection{Model Inversion Attack}
\label{sec:model_inversion}

Model inversion attacks reconstruct sensitive input attributes by exploiting model outputs such as prediction probabilities or confidence scores. These attacks can succeed even with partial black-box access, enabling inference of protected attributes including medical conditions, device usage, or household presence ~\cite{he2019model, zhang2020secret, zhang2022model, nguyen2023re}. They are especially dangerous when models overfit or exhibit high confidence on training samples; for example, inversion on softmax outputs can recover an average face representation 
for a target class.

\noindent\textbf{Defences.} Enforcing black-box access that exposes only class labels rather than probabilities limits inversion feasibility \cite{he2019model}. Confidence obfuscation, output sanitisation, and 
\ac{DP} training suppress the overconfident predictions that fuel 
inversion~\cite{he2019model}.

\subsection{Reconstruction Attack}
\label{sec:reconstruction_attack}

Reconstruction attacks recover original inputs from intermediate model representations such as latent feature vectors. By exploiting stored or exposed embeddings, adversaries can reconstruct raw \ac{IoT} inputs including sensor readings and images, compromising user confidentiality \cite{hilprecht2019monte, salem2020updates, zhao2024loki, 
Pham2025RAIFLE, gong2026FL}. An attacker with access to these feature vectors can reconstruct substantial portions of the original training data, threatening data confidentiality at scale.

\noindent\textbf{Defences.} Avoiding storage of explicit feature 
embeddings~\cite{hilprecht2019monte}, restricting access to internal parameters, and applying differentially private training or embedding obfuscation distort latent features sufficiently to make accurate recovery infeasible \cite{salem2020updates, chen2022practical}.

\subsection{Gradient Leakage Attack}
\label{sec:gradleak}

Gradient leakage attacks exploit gradients exchanged during distributed training to reconstruct original inputs. Unlike model inversion, which relies on model outputs, gradient leakage enables precise input recovery even without prediction scores. Introduced by Zhu \textit{et al.}~\cite{zhu2019deep}, these attacks pose a critical 
privacy risk in \ac{FL}~\cite{fan-boosting2025, fan2025refiner} and \ac{IoT} systems where gradients are frequently shared between devices and servers. The attack solves an optimisation problem searching for an input--label pair whose gradient best matches the observed client gradient. Techniques such as iDLG~\cite{zhao2020idlg} improve 
convergence and reconstruction quality, enabling data extraction from a single shared gradient step. This threat is particularly acute in \ac{IoT}-driven \ac{FL}, where constrained edge devices may inadvertently leak sensitive sensor or biometric information through gradient updates despite never transmitting raw data.

\noindent\textbf{Defences.} Gradient clipping and noise injection limit inversion precision \cite{abadi2016deep, sun2019can}. Secure aggregation prevents direct exposure of individual gradients \cite{bonawitz2017practical}, and larger batch sizes with shuffling introduce ambiguity in gradient origins~\cite{geiping2020inverting}.

\subsection{Adversarial Attack}
\label{sec:adv_attack}

Adversarial attacks involve deliberate manipulation of input data through carefully crafted perturbations designed to mislead \ac{ML} models \cite{goodfellow2014explaining}. 
These perturbations are often imperceptible to humans yet cause significant degradation in model performance. Although primarily threats to model integrity and availability, adversarial attacks intersect with privacy concerns by enabling inference about training data. The widely studied \ac{FGSM}~\cite{goodfellow2014explaining} perturbs inputs along the gradient direction to maximise prediction error.

In \ac{IoT} applications such as electric load forecasting \cite{almalaq2017review}, adversarial perturbations can produce inaccurate demand predictions~\cite{chen2019exploiting}, causing resource misallocation or grid-level instabilities. Their relevance to \ac{PPML} stems from the shared requirement for resilient, trustworthy models in environments handling sensitive or critical information.

\noindent\textbf{Defences.} Adversarial training is among the most effective approaches \cite{madry2017towards, chakraborty2021survey, costa2024survey}, complemented by gradient masking, input preprocessing, and certified defences offering formal robustness guarantees. Many techniques carry significant computational overhead or fail under adaptive adversaries \cite{carlini2017towards}, leaving resilient defence for \ac{IoT}-based \ac{PPML} an open research direction.

\subsection{Traffic Analysis Attack}
\label{sec:traffic_analysis}

Traffic analysis attacks infer sensitive information from communication patterns rather than model outputs or parameters. Even when model updates and sensor data are encrypted, observable metadata---packet timing, payload sizes, transmission frequency, and update 
schedules---can reveal user activity patterns, occupancy schedules, or device configurations~\cite{msadek2019iot, ding2022adversarial}. In \ac{IoT}-\ac{FL} deployments, gradient exchange traffic is particularly vulnerable: the volume and cadence of model updates can identify which devices are active, correlate updates with real-world events, and enable side-channel reconstruction of sensitive behavioural patterns. This threat is amplified in bandwidth-constrained networks such as 
\ac{LPWAN}s, where traffic sparsity makes individual transmissions more identifiable.

\noindent\textbf{Defences.} \ac{LDP} applied to transmitted updates adds noise prior to transmission, obscuring update content \cite{mahawaga2022local}. Communication scheduling and traffic 
shaping---padding transmissions to uniform sizes and randomising 
timing---reduce the informativeness of observable metadata~\cite{senol2026FLlora}. In 5G/6G \ac{IoT} architectures, slice-level resource isolation further limits cross-slice traffic inference~\cite{Zhao2025, zhang2025FL6G}.

\section{Evaluation Metrics in PPML}
\label{sec:evaluation_metric}

Evaluation in \ac{PPML} requires simultaneous assessment of privacy guarantees and model utility, given the inherent privacy--utility trade-off. Existing approaches can be grouped into three complementary categories: (i) formal privacy guarantees, (ii) empirical attack-based evaluation, and (iii) robustness analysis. A persistent limitation across the surveyed literature is that these categories are rarely reported together: most works optimise for one dimension while treating others as secondary. Rigorous \ac{PPML} evaluation requires joint reporting across all three, particularly in \ac{IoT} settings where system-level constraints interact with privacy parameters in ways that single-metric reporting cannot capture \cite{ponomareva2023dp, zaman2024privacy}.

\subsection{Privacy Metrics}

\subsubsection{Formal Guarantees}

The standard formal metric is the \ac{DP} privacy budget $\epsilon$, which bounds the distinguishability between neighbouring datasets. Smaller $\epsilon$ implies stronger privacy but typically reduces model accuracy. In practice, acceptable $\epsilon$ values are application-dependent, ranging from ${<}1$ in statistical analysis to ${\leq}10$ in \ac{DL} settings \cite{ponomareva2023dp}. The effective privacy--utility balance further depends on dataset scale and model capacity. For \ac{FL} systems, privacy accounting must additionally track composition across rounds: per-round 
$\epsilon$ values underestimate cumulative exposure in long-lived deployments, motivating the use of R\'{e}nyi \ac{DP} \cite{mironov2017renyi} accountants and zero-concentrated \ac{DP} for tighter composition bounds \cite{zheng2025RelaxedDP}.

\subsubsection{Empirical Privacy Risk}
Since formal guarantees may not reflect practical leakage, empirical evaluation against \ac{MIA}s~\cite{shokri2017membership} is widely adopted. Common indicators include: (i) \ac{MIA} accuracy, (ii) attack success rate (ASR), and (iii) \ac{ROC}--\ac{AUC}, where an \ac{AUC} of 0.5 denotes random guessing and higher values indicate increased leakage \cite{zaman2024privacy}. These metrics quantify adversarial inference capability under realistic threat models. For generative models, reconstruction-based metrics, such as the ability of an adversary to recover training samples from synthetic outputs, provide complementary leakage signals \cite{hilprecht2019monte}. Combining formal and empirical metrics is particularly important when $\epsilon$ is large or loosely calibrated, as formal guarantees in such regimes provide limited practical assurance.

\subsubsection{Robustness-Linked Metrics}
Adversarial robustness is increasingly considered a proxy for privacy 
sensitivity \cite{tramer2022detecting}. Metrics such as adversarial success rate (e.g., under \ac{FGSM}~\cite{goodfellow2014explaining}) and perturbation norms assess model sensitivity to input manipulations. High sensitivity may correlate with exploitable information leakage. In \ac{IoT} contexts, robustness evaluation should additionally account for sensor noise and distribution shift across devices, as perturbation norms calibrated for clean benchmark data may not transfer to heterogeneous field conditions.

\subsection{Utility Metrics}
Utility evaluation measures predictive or generative performance under privacy constraints. For classification, accuracy and F1-score remain standard, with F1 preferred under class imbalance. For regression and forecasting, \ac{RMSE} and \ac{MAE} quantify deviation from ground truth, with \ac{RMSE} penalising large errors more heavily.

For synthetic data in \ac{PPML}, evaluation focuses on distributional fidelity, temporal consistency, and downstream task transferability. Widely used metrics include \ac{KL} divergence and \ac{MMD} for statistical similarity~\cite{kullback1951information, gretton2012kernel}, \ac{DTW} for temporal alignment~\cite{berndt1994using}, and \ac{TSTR} for task-level generalisation~\cite{esteban2017real}. \ac{TSTR} is particularly valuable in \ac{IoT} settings because it directly measures whether synthetic data can substitute for real data in downstream tasks such as anomaly detection or load forecasting, without requiring access to real test samples during evaluation.

Effective benchmarking in \ac{PPML} requires joint reporting of privacy parameters, empirical leakage measures, and task-level utility. Table \ref{tab:eval_summary} summarises the recommended metric combinations by paradigm and task type.

\begin{table}[ht]
\centering
\caption{Recommended evaluation metrics by PPML paradigm and task type.}
\label{tab:eval_summary}
\small
\begin{tabular}{p{3.0cm}|p{3.5cm}|p{3.5cm}|p{3.5cm}}
\hline
\textbf{Paradigm} & \textbf{Privacy} & \textbf{Utility} & \textbf{Robustness} \\
\hline
Perturbation-Based (\ac{DP}) &
$\epsilon$, $\delta$; \ac{MIA} \ac{AUC} &
Accuracy; \ac{RMSE}; \ac{MAE} &
\ac{FGSM} ASR; perturbation norm \\
\hline
Distributed Learning (\ac{FL}) &
Per-round $\epsilon$; cumulative \ac{DP} budget; \ac{MIA} ASR &
Convergence rate; accuracy under \ac{non-IID} &
Byzantine resilience; gradient norm \\
\hline
Cryptographic (\ac{HE}/\ac{SMPC}) &
Formal security parameter; leakage probability &
Latency; throughput; \ac{RMSE} &
Side-channel resistance \\
\hline
Generative (\ac{GAN}) &
\ac{MIA} \ac{AUC}; reconstruction risk &
\ac{KL}; \ac{MMD}; \ac{DTW}; \ac{TSTR} &
Mode collapse rate; \ac{FGSM} ASR \\
\hline
\end{tabular}
\end{table}

\section{Datasets}
\label{sec:datasets_lit}

Publicly available datasets used in this survey span two dominant \ac{IoT} sensing modalities: visual observation and continuous monitoring. Image datasets illustrate privacy risks in camera-enabled \ac{IoT} systems, while electricity time-series provide established benchmarks for evaluating privacy--utility trade-offs in sequential sensor streams. A notable limitation of the current \ac{PPML}-\ac{IoT} literature is its concentration on these two modalities: mobility traces, audio streams, and industrial sensor data, all common in real \ac{IoT} deployments, remain underrepresented in privacy evaluation benchmarks, representing an open gap 
for the community.

\subsection{Image Datasets}
\label{sec:image_data}
\textit{MNIST}~\cite{mnist} is the standard handwritten digit benchmark 
of the same dimensions and is commonly used as a low-complexity baseline for comparing privacy--utility trade-offs across \ac{PPML} paradigms. 
\textit{Fashion-MNIST}~\cite{xiao2017fashion} contains 70,000 grayscale 
$28{\times}28$ images across 10 clothing categories (60,000 train / 10,000 test). It is widely used to benchmark \ac{DP-SGD} and \ac{MIA} defences under controlled conditions. 
\textit{CIFAR-10} ~\cite{cifar10} provides 60,000 colour $32{\times}32$ images across 10 object categories (50,000 train / 10,000 test) and is the standard benchmark for evaluating \ac{DP} at higher model complexity. 
\textit{\ac{LFW}}~\cite{huang2008labeled} comprises over 13,000 web-collected facial images labelled by identity and is specifically used to evaluate model inversion and reconstruction attack defences in face recognition pipelines. 
The \textit{Gender Identification} dataset~\cite{genderKeras} contains approximately 2,300 cropped facial images labelled male or female, used in attribute inference evaluations.

\subsection{Electricity Consumption Datasets}
\label{sec:time_data}

\textit{Pecan Street Dataport} \cite{street2019dataport} provides circuit-level consumption recorded at 15-minute intervals throughout 2018, capturing both consumption and grid-contribution values. Its appliance-level granularity makes it particularly suited to evaluating \ac{DP} mechanisms against load disaggregation and occupancy inference attacks. 
\textit{\ac{ECL}}~\cite{trindade2015ecl} records 15-minute electricity consumption from 370 Portuguese clients between 2011 and 2014, 
commonly resampled to hourly frequency and reduced to approximately 320 clients after preprocessing for sparsity. It is widely used for long-term forecasting and privacy--utility evaluation in sequential \ac{IoT} data. 
\textit{\ac{NREL}} ~\cite{muratori2017impact} covers 200 Midwestern US households sampled every 10 minutes (144 samples/household/day), capturing diverse usage patterns across varying household sizes and occupancy levels, and is used to evaluate synthetic data generation fidelity under \ac{DP} constraints.

\section{Evaluation Frameworks}
\label{sec:ppml_framework}

Several open-source frameworks support the development and empirical evaluation of 
\ac{PPML} techniques. Table~\ref{tab:frameworks} provides a structured comparison 
across the dimensions most relevant to \ac{IoT} deployment.

\begin{table}[ht]
\centering
\caption{Comparison of open-source PPML evaluation frameworks.}
\label{tab:frameworks}
\small
\begin{tabular}{p{2.0cm}|p{2.2cm}|p{2.6cm}|p{2.2cm}|p{2.4cm}|p{2.4cm}}
\hline
\textbf{Framework} & \textbf{Backend} & \textbf{DP Support} & 
\textbf{FL Support} & \textbf{Privacy Accounting} & \textbf{IoT Suitability} \\
\hline
TensorFlow Privacy~\cite{abadi2016deep} &
TensorFlow &
\ac{DP-SGD}, Gaussian, Laplace &
Via TFF &
R\'{e}nyi \ac{DP}; $(\epsilon,\delta)$ &
Moderate --- TF Lite compatible \\
\hline
Opacus~\cite{yousefpour2021opacus} &
PyTorch &
\ac{DP-SGD}; per-sample gradients &
Limited &
R\'{e}nyi \ac{DP}; PRV accountant &
Moderate --- edge PyTorch deployments \\
\hline
Flower (FLwr)~\cite{beutel2020flower} &
Agnostic (TF, PyTorch, others) &
Via integration with Opacus/TFP &
Native; heterogeneous devices &
Via external accountants &
High --- designed for edge/embedded \\
\hline
PySyft~\cite{ziller2021pysyft} &
PyTorch &
\ac{DP} via integration &
Native \ac{SMPC}; \ac{FL} &
Limited built-in &
Low --- high compute requirements \\
\hline
FATE~\cite{liu2021fate} &
Standalone &
\ac{DP} via integration &
Native; cross-silo \ac{FL} &
Limited built-in &
Low --- enterprise-oriented \\
\hline
\end{tabular}
\end{table}

\textbf{TensorFlow Privacy}~\cite{abadi2016deep} extends TensorFlow with 
\ac{DP}-aware training tools, including \ac{DP-SGD} and R\'{e}nyi \ac{DP} accounting for tracking cumulative $(\epsilon,\delta)$ loss across iterations. Its compatibility with TensorFlow Lite makes it the most accessible option for constrained \ac{IoT} inference pipelines.

\textbf{Opacus}~\cite{yousefpour2021opacus} integrates \ac{DP} into PyTorch workflows with efficient per-sample gradient computation via a hooks-based implementation. It supports the PRV (Privacy Random Variables) accountant for tighter composition bounds than standard R\'{e}nyi \ac{DP}, making it well-suited for iterative training under strict privacy budgets.

\textbf{Flower (FLwr)}~\cite{beutel2020flower} is a framework-agnostic \ac{FL} platform supporting TensorFlow, PyTorch, and other \ac{ML} backends. Its client--server architecture is explicitly designed for heterogeneous and resource-constrained devices, with built-in support for \ac{DP} via Opacus or TensorFlow Privacy integration and secure aggregation. Of the frameworks surveyed, Flower most directly addresses \ac{IoT} deployment realities.

\textbf{PySyft}~\cite{ziller2021pysyft} extends PyTorch with \ac{SMPC} and \ac{FL} primitives, enabling privacy-preserving computation across untrusted parties. While powerful for cryptographic \ac{PPML} research, its computational overhead makes direct \ac{IoT} deployment impractical without significant optimisation.

\textbf{FATE} (Federated AI Technology Enabler)~\cite{liu2021fate} is an 
enterprise-oriented \ac{FL} framework supporting cross-silo federation with integrated \ac{HE} and \ac{SMPC} components. Its architecture targets institutional deployments rather than resource-constrained edge systems, limiting its direct applicability to \ac{IoT} settings.

Collectively, these frameworks underpin reproducible \ac{PPML} research. For \ac{IoT}-focused work, Flower combined with Opacus or TensorFlow Privacy currently represents the most practical stack, balancing \ac{FL} flexibility, \ac{DP} accounting rigour, and edge deployment compatibility. A gap remains for frameworks natively supporting \ac{DP-BF}-Encoding, TinyML-compatible privacy accounting, and end-to-end evaluation pipelines spanning sensing, communication, and inference 
stages.

\section{Open Issues, Opportunities, and Future Trends in PPML for IoT}
\label{sec:future}

Although \ac{PPML} has advanced significantly, a number of challenges remain for deployment in resource-constrained and heterogeneous \ac{IoT} environments. This section synthesises open issues, emerging opportunities, and future research trends 
across privacy mechanisms, model training paradigms, emerging architectures, and deployment governance.

\subsection{Privacy Mechanism Challenges in IoT}

\subsubsection{DP-Based Mechanisms: Challenges Across Deployment Contexts}
\label{sec:dp_challenges_future}

\ac{DP} provides strong formal guarantees but faces distinct challenges at each stage of the \ac{IoT} pipeline. At the \textit{input level}, local perturbation requires careful 
calibration to balance utility against energy, computation, and communication costs; systematic approaches tailored to \ac{IoT}-specific workloads remain limited despite recent adaptive noise allocation strategies~\cite{liu2024multi}. At the \textit{training level}, \ac{DP-SGD} assumes access to raw or minimally processed data, conflicting with decentralised \ac{IoT} scenarios where data ownership and regulatory constraints prohibit aggregation. At the \textit{prediction level}, output perturbation incurs minimal training overhead but does not protect model parameters or intermediate 
representations, limiting its effectiveness in long-lived \ac{IoT} inference pipelines.

When extended to \textit{synthetic data generation}, \ac{DP} introduces additional tensions. Differentially private variants such as \ac{DP-GAN}~\cite{xie2018differentially, torkzadehmahani2019dp} and PATE-GAN~\cite{jordon2018pate} provide formal guarantees but degrade temporal fidelity, which 
is critical for \ac{IoT} tasks such as energy forecasting, healthcare monitoring, and anomaly detection~\cite{zhang2018generative, fekri2019generating, li2024deep}. Generative models also remain prone to mode collapse, suppressing minority patterns that correspond to rare but operationally critical \ac{IoT} events \cite{zhang2023generative}. Furthermore, existing benchmarks primarily target visual or tabular datasets, providing limited insight into time-series or multi-modal \ac{IoT} data~\cite{golda2024privacy}.

\textit{Future research} should prioritise hybrid \ac{DP} frameworks combining input-level privatisation with lightweight encoding or secure aggregation, with adaptive privacy budgets that account for data sensitivity and device-level constraints. For synthetic data, hybrid generative architectures such as \ac{VAE}--\ac{GAN} \cite{razghandi2022variational} and transformer-based models \cite{li2022tts} offer improved temporal modelling under integrated \ac{DP} constraints. Integrated evaluation 
frameworks that jointly assess formal privacy guarantees, disclosure risk, and downstream utility are essential for practical \ac{IoT} adoption \cite{zaman2025synthetic}.

\subsubsection{Residual Challenges in Cryptographic PPML for IoT}
\label{sec:crypto_limitations}

Cryptographic \ac{PPML} techniques provide strong theoretical guarantees but face substantial deployment barriers in \ac{IoT} systems. \ac{HE} enables computation on encrypted data but introduces prohibitive latency from ciphertext expansion and homomorphic evaluation, with memory and energy footprints exceeding typical microcontroller-class devices \cite{cheon2017homomorphic, chillotti2020tfhe, 
zhang2023lightweight}. \ac{SMPC} eliminates reliance on trusted central authorities but exhibits poor scalability for \ac{DL}, with high communication overhead for non-linear 
operations in bandwidth-constrained \ac{IoT} networks \cite{boyle2019secure, liu2017oblivious, ma2024trusted}. \ac{TEE}s reduce cryptographic overhead through hardware-isolated enclaves but are constrained by limited secure memory, vendor trust dependencies, and susceptibility to side-channel and speculative execution attacks \cite{costan2016intel, schneider2022sok}.

\textit{Future research} should develop hybrid frameworks combining cryptographic protection with perturbation-based methods such as \ac{DP} to reduce overhead, harden \ac{TEE} deployments through runtime attestation and enclave monitoring, and integrate post-quantum cryptographic primitives for long-lived \ac{IoT} systems.

\subsubsection{Probabilistic Encoding with DP-BF for IoT}
\label{sec:dpbf_future}

\ac{DP-BF}-Encoding offers a lightweight privacy mechanism well-suited to resource-constrained \ac{IoT} devices \cite{wang2025DataEncode, ulfer2014, xue, 
zaman2024privacy}. However, compatibility with end-to-end deep \ac{DL} architectures remains limited, standardised benchmarks for comparing \ac{DP-BF} against alternative 
\ac{PPML} techniques are absent, and evaluations predominantly focus on smart metering with limited coverage of mobility, localisation, or ambient sensing domains. \textit{Future work} should develop adaptive encoding mechanisms that dynamically adjust the privacy budget ($\epsilon$) based on device capabilities and workload, integrate \ac{DP-BF} into 
\ac{FL} workflows to limit gradient reconstruction risk, and extend validation across diverse \ac{IoT} domains.

\subsection{Model Training Paradigms}

\subsubsection{Federated, Transfer, and Split Learning: Privacy and Deployment Challenges}
\label{sec:fl_sl_merged}

\ac{FL} preserves data locality but remains vulnerable to gradient leakage \cite{zhu2019deep}, and integrating \ac{DP} into \ac{FL} pipelines can significantly impair convergence under the \ac{non-IID} data conditions prevalent across 
heterogeneous \ac{IoT} devices. Uniform privacy budget allocation exacerbates this by ignoring differences in data sensitivity and device capability, while system heterogeneity complicates synchronisation and straggler management \cite{mcmahan2017communication}.

\ac{FTL} extends \ac{FL} by enabling knowledge reuse across tasks and domains without centralising data, but introduces additional privacy risks: transferred representations may encode sensitive patterns from source domains, and domain mismatch can reduce the usefulness of \ac{DP}-protected representations. Cold-start conditions further limit standard \ac{FL} effectiveness when devices lack sufficient local data.

\ac{SL}~\cite{yu2025Split} and its variants such as SplitFed~\cite{zhao2025SplitFed} reduce device-side computation by transmitting intermediate activations, but these activations can leak sensitive information depending on the split configuration \cite{Thapa_Mahawaga_Arachchige_Camtepe_Sun_2022, iqbal2025privacy}. Early split points 
expose features correlated with raw sensor data, while deeper splits shift computational burden back to resource-constrained devices. TinyML~\cite{kiran2025tinyML, banbury2021micronets, behera2025tinyML} enables on-device inference and minimises 
data transmission but lacks runtime support for adaptive privacy accounting, making static and overly conservative privacy configurations common.

\textit{Future research} should pursue unified edge-centric frameworks integrating \ac{FL}, \ac{SL}, and TinyML, with lightweight privacy mechanisms such as \ac{DP-BF}-Encoding or activation perturbation at split interfaces. Adaptive privacy budgeting, privacy-preserving knowledge distillation, and cross-layer optimisation jointly considering privacy, communication efficiency, and learning performance will be essential for scalable \ac{IoT} deployment \cite{tinyMLLamaakal2025}.

\subsection{Emerging Architectures and Analytics}

\subsubsection{Privacy-Aware Foundation Models and Multimodal Analytics for IoT}
\label{sec:foundation_multimodal_iot}

Foundation models, including \ac{LLM}s, vision-language models, and multimodal transformers, are increasingly adapted for \ac{IoT} analytics through numerical tokenisation, sequence prompting, and cross-modal representation learning \cite{dell2025agentic, jin2023large, wang2025IoT-LLM}. Their ability to generalise across tasks and domains with minimal labelled data is particularly valuable in \ac{IoT} environments where sensing configurations evolve continuously~\cite{xue2023utilizing, 
wu2024stellm, zhang2025soft}. However, fine-tuning on sensitive domain-specific data, capturing energy usage, mobility trajectories, or physiological signals, can implicitly encode personal patterns recoverable through membership inference, data extraction, or prompt-based attacks~\cite{fu2025membership, qu2025prompt}. Multimodal architectures amplify this risk by enabling cross-modal re-identification even when individual streams appear privacy-safe in isolation \cite{chen2025unveiling}. Applying \ac{DP} during fine-tuning degrades long-range temporal dependencies critical for \ac{IoT} time-series tasks, and the computational footprint of these models limits on-device deployment \cite{zhao2024privacy, zhang2025hfcc, liu2026LLM}.

Cross-domain generalisation further compounds these challenges: distributional shifts across \ac{IoT} domains amplify \ac{DP}-induced utility loss, and adversarial threat models differ across domains—biometric inference in healthcare versus load disaggregation in smart grids, complicating universal defence design. Naively reusing 
privacy budgets across tasks risks cumulative leakage, while domain adaptation can inadvertently expose sensitive features from source domains.

Future research should explore modality-specific sensitivity allocation within transformer architectures \cite{higashi2025PEFT}, combine personalised \ac{FL} with secure aggregation for distributed foundation model fine-tuning~\cite{wang2025FL}, establish systematic post-deployment auditing including cross-modal linkage probes 
and membership inference stress tests~\cite{wang2025mia}, and develop cross-domain privacy accounting mechanisms that track cumulative privacy loss during model reuse. 
\ac{IoT}-specific benchmarking suites spanning multiple tasks, modalities, and deployment tiers remain urgently needed.

\subsubsection{Quantum-Aware Privacy for IoT}
\label{sec:quantum_privacy}
\ac{QML} promises computational advantages for optimisation-heavy \ac{IoT} workloads but introduces distinct privacy risks \cite{wang2024QAI}. Most practical \ac{QML} systems rely on hybrid quantum--classical pipelines where intermediate states exchanged between classical and quantum components may expose sensitive representations. Classical \ac{DP} is analysed under classical adversarial assumptions, but quantum-capable adversaries may leverage search acceleration and information amplification to weaken existing protections, a critical 
concern for \ac{IoT} datasets requiring long-term confidentiality, such as healthcare records and smart grid measurements.

\ac{PQ-DP} has emerged to address this threat, but its deployment in resource-constrained \ac{IoT} systems remains challenging: lattice-based post-quantum primitives incur substantial memory and computational overhead unsuitable for microcontroller-class devices, and composition rules for classical--quantum hybrid pipelines remain unresolved \cite{watkins2023quantum, satpathy2023analysis}. \textit{Future research} should develop lightweight \ac{PQ-DP} mechanisms emphasising local perturbation at the data source, 
quantum-aware privacy accounting and composition rules, and integration of \ac{PQ-DP} with post-quantum secure \ac{FL} combining lattice-based encryption with \ac{DP} for end-to-end quantum-resilient \ac{PPML} pipelines.

\subsection{Deployment, Governance, and Lifecycle Challenges}
\label{sec:deployment_governance}

Translating \ac{PPML} into real-world \ac{IoT} deployments requires a shift from mechanism-centric design toward system-level governance. Most techniques are evaluated in isolation, but deployed systems involve multiple coexisting privacy mechanisms interacting across sensing, communication, learning, and inference stages—interactions that can introduce unintended leakage or utility degradation. Privacy budgets must be treated as managed system resources requiring runtime monitoring and policy-driven adaptation, with such adaptation remaining transparent and auditable to avoid introducing new side channels. As \ac{IoT} analytics increasingly fuse data across devices, modalities, and administrative domains, privacy loss can accumulate across tasks and time; governance frameworks must therefore track cumulative exposure at the system level rather than individual components.

Regulatory frameworks such as \ac{GDPR} and \ac{HIPAA} additionally impose accountability, transparency, and explainability requirements that many \ac{PPML} techniques lack built-in support for, necessitating logging, provenance tracking, and post-hoc analysis tools. Over longer timescales, models are retrained, devices replaced, and regulatory interpretations revised—requiring lifecycle-aware strategies for safe model updates, controlled budget renewal, and periodic privacy risk reassessment.

In next-generation \ac{IoT} systems operating over 5G and 6G infrastructures \cite{zhang2025FL6G}, privacy preservation becomes intrinsically a communications systems design problem. Over-the-Air Computation (AirComp) \cite{wang2024AirComp} enables efficient \ac{FL} aggregation \cite{ni2026FL} by exploiting wireless channel superposition, but injected \ac{DP} noise interacts with channel distortion and fading effects in ways that remain poorly characterised. Low-power wide-area networks (LPWAN) such as LoRaWAN and NB-IoT impose sub-kilobit-per-second uplinks and duty-cycle constraints incompatible with most existing \ac{PPML} frameworks, requiring joint optimisation of update compression, communication scheduling, and privacy calibration \cite{senol2026FLlora, dkhar2025LPWAN}. Additionally, \ac{PPML} traffic patterns—gradient exchanges and synthetic dataset dissemination, introduce identifiable signals that enable side-channel attacks, extending privacy requirements to slice-level resource isolation and traffic shaping in 5G/6G network slicing architectures \cite{Zhao2025}.

\textit{Future research} should develop integrated governance frameworks bridging algorithmic privacy with operational control, policy-aware privacy orchestration, system-level privacy accounting across components and time, and communication-theoretic privacy analysis for next-generation \ac{IoT} network architectures.

\subsection{Summary and Outlook}
\label{sec:ppml_summary}

Across all paradigms examined, a consistent theme emerges: effective privacy preservation in \ac{IoT} requires system-level thinking beyond isolated algorithmic guarantees. \ac{IoT} constraints on computation, energy, and communication fundamentally limit the feasibility of theoretically sound mechanisms on heterogeneous, resource-constrained devices. The increasing use of foundation models, multimodal analytics, and cross-domain learning introduces new risks—model memorisation, cross-modal inference, and cumulative privacy loss—that challenge existing threat models, further amplified by emerging quantum adversarial capabilities. The transition to real-world 
deployment reveals the central role of governance, accountability, and lifecycle management: privacy budgets must be auditable over time and aligned with evolving 
regulatory frameworks. Future work should prioritise holistic privacy architectures coordinating mechanisms across the data lifecycle, standardised benchmarks jointly evaluating privacy and utility under realistic conditions, and deployment-aware 
frameworks aligning technical guarantees with regulatory and societal expectations.

\subsection{Lessons Learned}

The cross-paradigm analysis presented in previous sections, together with the taxonomy in Fig.\ref{fig:taxonomy} and comparative summaries in Tables (\ref{tab:ppml_paradigm_comparison}, \ref{tab:ppml_comparison}), yields several system-level insights that inform the practical design of \ac{PPML} for \ac{IoT} ecosystems.

\textit{1) Cross-paradigm integration is necessary for robust IoT privacy.}
The taxonomy in Fig.\ref{fig:taxonomy} demonstrates that perturbation-based methods (\ac{DP}), distributed learning (\ac{FL}), cryptographic protocols, and generative models address distinct threat surfaces and deployment constraints. As shown in Table \ref{tab:ppml_paradigm_comparison}, no single paradigm simultaneously provides formal privacy guarantees, strong adversarial robustness, high utility retention, and lightweight \ac{IoT} deployability. \ac{DP} offers quantifiable $(\epsilon,\delta)$ guarantees but introduces utility degradation under tight budgets (Section \ref{sec:dp_lit}). \ac{FL} reduces centralised exposure but remains vulnerable to gradient leakage and Byzantine behaviour (Section \ref{sec:fl}, Table \ref{tab:ppml_comparison}). Cryptographic methods provide strong confidentiality independent of statistical assumptions, yet incur substantial computational and communication overhead (Section \ref{sec:cryptographic}). Generative models reduce raw data sharing but lack formal guarantees unless integrated with DP (Section \ref{sec:synthetic_data}). These observations suggest that hybrid architectures, such as \ac{DP}-enhanced \ac{FL}, secure aggregation with lightweight encryption, \ac{DP}-enhanced data encoding \cite{zaman2024privacy} or \ac{DP}-integrated generative synthesis \cite{zaman2025synthetic}, represent a necessary direction for scalable IoT deployment.

\textit{2) The privacy--utility--efficiency trade-off is amplified in IoT settings.}
Section \ref{sec:privacy_utility} established the inherent privacy--utility tension in ML systems. In IoT deployments, this tension extends into a three-way optimisation problem involving privacy guarantees, predictive performance, and system efficiency. Lower $\epsilon$ values in \ac{DP} increase noise magnitude; \ac{SMPC} increases communication rounds; \ac{HE} raises latency and memory costs; and \ac{FL} under heterogeneous, \ac{non-IID} sensor data may slow convergence (Sections \ref{sec:dp_lit}–\ref{sec:fl}). Because \ac{IoT} devices are resource-constrained and often latency-sensitive, privacy-preserving mechanisms must be evaluated not only for formal guarantees but also for energy consumption, bandwidth overhead, and convergence stability (Table \ref{tab:ppml_comparison}). Privacy in IoT cannot be treated as an add-on; it must be co-optimised with system constraints from the outset.

\textit{3) Threat models and defences must be jointly engineered.}
The attack taxonomy in Section \ref{sec:active_passive} highlights that privacy threats vary across white-box/black-box and passive/active adversaries. Section \ref{sec:ppml_attack} further shows that reconstruction, model inversion, gradient leakage, and membership inference attacks exploit different aspects of the learning pipeline. A defence effective against one threat class may leave exposure elsewhere, for example, \ac{DP} mitigates membership inference but does not inherently prevent model extraction or adversarial manipulation; secure aggregation protects against server-side inference but not malicious client updates. Therefore, privacy mechanisms must be selected in alignment with explicitly stated adversary models, rather than applied generically. Adversary-aware co-design is particularly critical in distributed \ac{IoT} environments where trust assumptions differ across devices, gateways, and cloud infrastructure.

\textit{4) Evaluation practices must reflect realistic IoT deployment conditions.}
Although Section \ref{sec:evaluation_metric} consolidates formal, empirical, and robustness-based evaluation metrics, a recurring pattern across the surveyed literature is reliance on standard image benchmarks or resource-rich GPU environments. However, Section \ref{sec:iot_privacy} emphasises that IoT systems operate under heterogeneous, bandwidth-limited, and intermittently connected conditions. Techniques validated on MNIST \cite{mnist} or CIFAR \cite{cifar10} like datasets may exhibit degraded performance when deployed on microcontroller-class hardware or under \ac{non-IID} sensor distributions. Future \ac{PPML} evaluation must therefore incorporate realistic \ac{IoT} workloads, communication constraints, and long-duration operational testing to bridge the gap between laboratory validation and field deployment.

\textit{5) Formal privacy guarantees and regulatory compliance are distinct objectives.}
The introduction situates \ac{IoT} privacy within regulatory frameworks such as \ac{GDPR}, \ac{HIPAA}, and \ac{PIPL}. However, as discussed throughout Sections \ref{sec:data_privacy} - \ref{sec:ppml_attack}, formal $(\epsilon,\delta)$-DP guarantees quantify statistical indistinguishability, whereas regulatory compliance concerns data minimisation, accountability, and purpose limitation. A \ac{DP}-trained model is not automatically regulation-compliant, and a regulation-compliant system may still leak through inference attacks. Effective \ac{PPML} deployment therefore requires alignment between formal privacy definitions, system-level safeguards, and governance mechanisms. Bridging this technical–legal gap remains an underexplored research direction.

\textit{6) Privacy guarantees must be analysed over system lifecycles, not single training rounds.}
Section \ref{sec:core_ppml} introduced \ac{DP} and its $(\epsilon,\delta)$ guarantees, which compose under repeated queries and retraining. In continuous \ac{IoT} deployments, where models are periodically retrained, federated rounds execute iteratively, and streaming data generates repeated queries, privacy loss accumulates according to composition properties. Reporting per-round $\epsilon$ values without accounting for long-term aggregation may substantially underestimate effective lifetime exposure. Sustainable \ac{IoT} \ac{PPML} therefore requires lifecycle-aware privacy accounting, including composition-aware privacy accountants, global budget tracking, and explicit retraining policies. Without such mechanisms, cumulative privacy leakage may silently exceed acceptable bounds in long-lived deployments.

\textit{7) Communication overhead is a first-order privacy–performance bottleneck.}
Section \ref{sec:fl} shows that in distributed \ac{PPML} systems, communication of gradients, model updates, or encrypted ciphertexts frequently dominates computational cost. In bandwidth-constrained \ac{IoT} networks, this overhead directly affects convergence time, energy consumption, and privacy budget utilisation. Techniques such as compression, sparsification, and quantisation reduce communication load but interact non-trivially with \ac{DP} noise injection and secure aggregation protocols. Joint optimisation of communication efficiency and privacy parameters is therefore essential for practical \ac{IoT}-scale \ac{PPML}.

Collectively, these lessons underscore that trustworthy \ac{PPML} for \ac{IoT} requires cross-paradigm integration, adversary-aware system design, lifecycle-conscious privacy accounting, and evaluation methodologies grounded in realistic deployment constraints. Privacy in \ac{IoT} is not merely a property of algorithms; it is an emergent characteristic of end-to-end system design.

\section{Conclusion}
\label{sec:conclusion}

\ac{PPML} has evolved into a critical research frontier for \ac{IoT} ecosystems, where data misuse risks intersect with constraints in compute, bandwidth, and deployment heterogeneity. This survey has presented a structured, cross-paradigm analysis spanning perturbation-based mechanisms (\ac{DP} and \ac{DP-BF}-Encoding), distributed learning (\ac{FL} and its hybrid variants), cryptographic protocols (\ac{HE}, \ac{SMPC}, \ac{TEE}s), and generative synthesis (\ac{DP-GAN}, Transformer-based \ac{GAN}s). For each paradigm, we examined formal privacy guarantees, computational trade-offs, adversarial robustness, and deployability under \ac{IoT} resource constraints.
A central finding is that no single paradigm simultaneously satisfies formal guarantees, adversarial robustness, high utility, and lightweight deployability. Hybrid architectures combining \ac{DP}-enhanced \ac{FL} with secure aggregation, lightweight encoding, or differentially private generative synthesis represent a necessary direction. The survey further mapped major attack vectors, including \ac{MIA}s, gradient leakage, model inversion, and reconstruction attacks, to corresponding defences, and consolidated evaluation metrics, benchmark datasets, and open-source frameworks to support reproducible research.
Looking ahead, priority challenges include energy-aware and TinyML-compatible privacy schemes, privacy-preserving \ac{LLM} fine-tuning under \ac{non-IID} \ac{IoT} distributions, \ac{PQ} \ac{DP} mechanisms, lifecycle-aware privacy accounting across federated rounds and retraining cycles, and governance frameworks bridging algorithmic guarantees with the accountability requirements of \ac{GDPR}, \ac{HIPAA}, and \ac{PIPL}. We advocate for a privacy-by-design mindset embedded throughout the \ac{ML} lifecycle, where data protection is a co-designed, auditable, and regulation-aligned cornerstone of scalable, trustworthy \ac{IoT} systems.


\bibliographystyle{ACM-Reference-Format}
\bibliography{reference}


\end{document}